\documentclass[letterpaper, 10 pt, journal, twoside]{IEEEtran}

\usepackage{cite}

   \usepackage[pdftex]{graphicx}
\usepackage{amsmath}
\usepackage{amssymb}
\usepackage{array}

 \usepackage[caption=false,font=footnotesize]{subfig}
\usepackage[ruled, linesnumbered]{algorithm2e}
\usepackage[dvipsnames]{xcolor}
\usepackage{pgfplots}
\pgfplotsset{compat=1.17}
\usepgfplotslibrary{fillbetween}
\usepackage{tikz}
\usepackage{float}
\usepackage{makecell}
\usepackage{comment}
\usepackage{upgreek}
\usepackage[normalem]{ulem}
\usepackage{url}

\definecolor{tumBlue}{HTML}{0065bd}
\definecolor{tumLighterBlue}{HTML}{98c6ea}
\definecolor{tumOrange}{HTML}{e37222}
\definecolor{tumGreen}{HTML}{a2ad00}
\definecolor{tumYellow}{HTML}{fed702}
\definecolor{tumPink}{HTML}{b55ca5}
\definecolor{tumDarkerBlue}{HTML}{072140}

\newcommand{\Korbinian}[1]{\textcolor{ForestGreen}{ (#1) }}
\newcommand{\Max}[1]{\textcolor{blue}{ (#1) }}
\newcommand{\David}[1]{\textcolor{red}{ (#1) }}

\newtheorem{assumption}{Assumption}

\newcommand{\T}{
	^{\top}
}

\newcommand{\q}{
    \mathbf{q}
}

\newcommand{\dq}{
    \Dot{\mathbf{q}}
}

\newcommand{\x}{
    \mathbf{x}
}

\newcommand{\bxi}{
    \boldsymbol{\xi}
}

\newcommand{\br}{
    \mathbf{r}
}

\newcommand{\bzeta}{
    \boldsymbol{\zeta}
}

\newcommand{\bpsi}{
    \boldsymbol{\psi}
}

\newcommand{\bu}{
    \mathbf{u}
}

\newcommand{\ba}{
    \mathbf{a}
}

\newcommand{\blambda}{
    \boldsymbol{\lambda}
}

\let\FINALVERSION=1

\ifx\FINALVERSION 0
    
    \newcommand{\rev}[1]{\textbf{\color{blue}#1}}
    \newcommand{\revMath}[1]{\color{blue}#1}
    \colorlet{RevColor}{blue}
    
    \newcommand{\revrm}[1]{\rev{\sout{#1}}}
    \newcommand{\revrmfig}[1]{{\color{blue}#1}}
\else
    \newcommand{\rev}[1]{{#1}}
    \newcommand{\revMath}[1]{#1}
    
    \newcommand{\revrm}[1]{\ignorespaces}
    \newcommand{\revrmfig}[1]{}
    \colorlet{RevColor}{black}
\fi

\def\thickhline{\noalign{\hrule height.8pt}}

\usepackage{subfiles}

\begin{document}

\markboth{IEEE Robotics and Automation Letters. Preprint Version. Accepted November, 2025}
{Griesbauer \MakeLowercase{\textit{et al.}}: Discovering Optimal Natural Gaits of Dissipative Systems}
%
\title{Discovering Optimal Natural Gaits of Dissipative Systems\\via Virtual Energy Injection}

%
%
%

\author{Korbinian~Griesbauer$^{1}$,
        Davide~Calzolari$^{1}$,
        Maximilian~Raff$^{2}$,
        C.~David~Remy$^{2}$,
        Alin~Albu-Sch\"{a}ffer$^{1}$
\thanks{Manuscript received: June 4, 2025; Revised October 2, 2025; Accepted November 10, 2025. This paper was recommended for publication by Editor Olivier Stasse upon evaluation of the Associate Editor and Reviewers' comments.
This work was supported by the European Research Council (ERC) through ERC Advanced Grant M-Runners under Grant 835284.}
\thanks{$^{1}$K. Griesbauer, D. Calzolari, and A. Albu-Sch\"{a}ffer are with School of Computation, Information and Technology, Technical University of Munich (TUM), 85748 Garching, Germany, and with the Institute of Robotics and Mechatronics, German Aerospace Center (DLR), 82234 Weßling, Germany {\fontfamily{qcr}\selectfont \{korbinian.griesbauer,davide.calzolari,alin.albu-schaeffer\}@dlr.de}}
\thanks{$^{2}$M. Raff and C.~D. Remy are with the Institute for Adaptive Mechanical Systems, University of Stuttgart, 70569 Stuttgart, Germany {\tt\footnotesize \{maximilian.raff,david.remy\}@iams.uni-stuttgart.de}}
\thanks{\copyright 2025 IEEE. Personal use of this material is permitted. Permission from IEEE must be obtained for all other uses, in any current or future media, including reprinting/republishing this material for advertising or promotional purposes, creating new collective works, for resale or redistribution to servers or lists, or reuse of any copyrighted component of this work in other works.}
}

\maketitle

\begin{abstract}
Legged robots offer several advantages when navigating unstructured environments, but they often fall short of the efficiency achieved by wheeled robots. One promising strategy to improve their energy economy is to leverage their natural (unactuated) dynamics using elastic elements. This work explores that concept by designing energy-optimal control inputs through a unified, multi-stage framework. It starts with a novel energy injection technique to identify passive motion patterns by harnessing the system’s natural dynamics.
This enables the discovery of passive solutions even in systems with energy dissipation caused by factors such as friction or plastic collisions. 
Building on these passive solutions, we then employ a continuation approach to derive energy-optimal control inputs for the fully actuated, dissipative robotic system. The method is tested on \rev{simulated} models to demonstrate its applicability in \revrm{different legged}\rev{both single- and multi-legged} robotic systems. This analysis provides valuable insights into the design and operation of elastic legged robots, offering pathways to improve their efficiency and adaptability by exploiting the natural system dynamics.
\end{abstract}


%

%
%
%
%

\section{Introduction}
\IEEEPARstart{T}{he} field of robotics is undergoing rapid advancements, with a particular emphasis on the development of humanoid robots that have gathered significant public attention. A key area of research focuses on developing robots capable of fast, safe, and efficient locomotion, modeled after human-like legs. A variety of possible motions enables such systems to move themselves forward. These periodic motions are called \textit{gaits}. However, an essential concern is achieving economic movement. This goal is typically pursued through the use of optimal control \rev{\cite{Wensing2024} or more specifically through methods like Differential Dynamic Programming\cite{li2020hybrid}.} \revrm{, which has been established as the standard tool for gait generation [1].}\par
\begin{figure}
    \centering
    \resizebox{0.98\linewidth}{!}{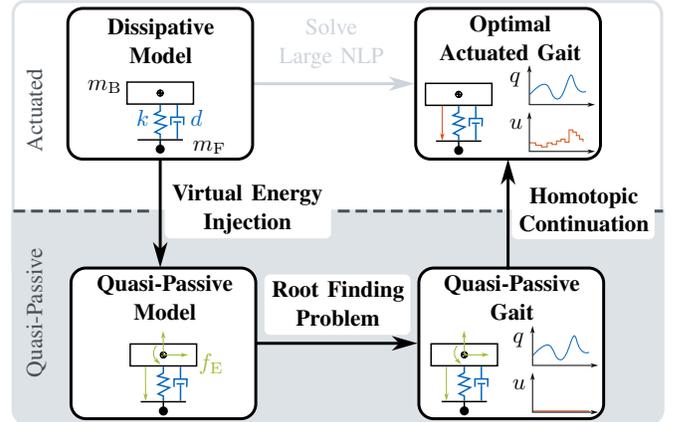}
    \caption{The presented approach utilizes three main concepts. We first apply \textit{virtual energy injection} to obtain a quasi-passive model. A quasi-passive system is not supported by actuation but through a non-physical energy input scaled by one scalar parameter. Through \textit{root-search}, we find a quasi-passive gait. The \textit{homotopic continuation} allows the derivation of a fully actuated optimal gait. Overall, this approach leverages the natural dynamics of the robotic system, thereby avoiding the need to solve large \revrm{nonlinear problems}\rev{NLPs} with high-dimensional decision variables.}
    \label{fig:Concept}
\end{figure}
Still, several concepts can be investigated to increase the energy economy of the systems themselves and simplify the search process. One such concept is the application of elasticity within the system \cite{xi2016selecting, werner2017generation}. Similarly to tendons in animals' legs, the springs can store elastic energy during the stance phase of running and release it during lift-off \cite{alexander2003principles}. This allows the exploitation of nonlinear modes to execute highly efficient motions \cite{calzolari2022single, della2021using}. The elasticity is especially advantageous for legged systems, as the springs can absorb impacts to protect the motors. Furthermore, springs enable motions that are not supported by any actuation, which we call \textit{passive}. Finding passive gaits is typically achieved by simplifying the system to exclude damping. The absence of damping enables the conservation of energy and the establishment of periodic motions, which are similar to the nonlinear modes of the mechanical system, without applying actuation~\cite{gan2018all}. An additional model simplification to sustain conservativeness is to take the limit of the foot mass going against zero $(m_\mathrm{F} \rightarrow 0)$, such that no energy is lost when the foot's velocity is brought to zero at impact~\cite{garcia1998simplest}. Conservative models also serve as ideal templates for generating energetically optimal gaits, as they enable locomotion with zero energy input \cite{full1999templates}. However, the modeling process leading to energy conservation significantly influences the system dynamics, posing the question of how to adapt the gaits found for the conservative system to the real robot. In particular, the need for vanishing damping can introduce unwanted oscillations, while the requirement for zero collision mass may alter the system’s degrees of freedom~\cite{Johnson2016}.\par
The standard approach for exploiting the elasticity of a system remains optimal control \cite{Wensing2024}. However, there are several challenges associated with its implementation. The search space for a nonlinear \rev{program (NLP)} increases drastically with the complexity of the system. This is due to the fact that the system states, as well as the actuation states, must be explored, resulting in a high-dimensional decision variable.
Moreover, for the solver to produce reliable results, the initial guess provided by the user must be reasonably close to the actual solution. The identification of entire families of gaits~\cite{raff2022connecting} in advance may be a prerequisite in this case. \rev{This can be bypassed by a modular approach.}\par%
In this paper, we seek to bridge between the passive template models and active optimal control. This will allow us to apply successful tools that exist for the analysis and generation of gaits in passive models \cite{remy2010stability, remy2011matlab} to a larger class of hybrid dynamic systems. The key concept that we introduce is \textit{quasi-passive models}. In such models, an energy source that is parameterized with a single parameter~$\gamma$ counteracts damping, friction, and other losses. This energy source does not necessarily have to be physically realizable and should not be thought of as a controller. Its only purpose is to counteract energy dissipation to enable periodic locomotion. In a subsequent continuously-parameterized optimization process, this input is gradually reduced and replaced by an actuation profile, which is physically feasible. Through this approach, we are able to leverage the natural dynamics of the system for efficient gait generation. The concept of quasi-passive models includes passive dynamic walkers on a slope \cite{mcgeer1990passive, rosa2023approach}. In these systems, the energy source is provided by the slope~$\gamma$ since it defines an external force (gravity) pulling the robot down the ramp.\par
This work can be seen as a generalization of \cite{rosa2023approach}, where families of gaits of zero actuation under the influence of gravity are found and connected to their optimally actuated counterparts. Similarly, Raff et al. \cite{raff2022generating} employ a simplified energetically conservative model to identify optimally actuated gaits through a model homotopy and numerical continuation on the optimality. The methods presented in \cite{raff2022generating} lay the foundation for our conceptual framework. Continuation is also useful to find entire families of gaits along solution manifolds, a process that is typically accomplished using energetically conservative models \cite{raff2022connecting, boffa2021quadrupedal}. The novel energy input introduced in our work is motivated by the periodic motion concept in \cite{krack2015nonlinear}, where the influence of damping onto nonlinear modes of a system is discussed.
\par
\rev{We contribute to the state of the art in efficient locomotion with elastic legged systems with the following:
\begin{enumerate}
    \item A novel strategy to counteract energy losses in elastic locomotion using a virtual input, which is parameterized with a single parameter (Sec.~\ref{sec:QuasiPassivity}).
    \item A modular approach to find optimally actuated gaits for elastic legged models including damping and other losses. The framework consists of a root-search with direct collocation and a homotopic continuation on the first-order optimality (Sec.~\ref{sec:Implementation}).
    \item A proof of concept simulated on two legged systems. The examples include a one- and a two-legged model with different actuation (Sec.~\ref{sec:Prism} and~\ref{sec:Quadru}).
\end{enumerate}
}
\revrm{In this paper, we first formulate the problem statement and introduce the underlying concepts to understand hybrid legged systems and their gaits (Sec.~II). We then focus on passive and quasi-passive gaits (Sec.~III), which constitute the foundation for homotopic continuation and the derivation of optimally actuated gaits (Sec.~IV). A description of the algorithms we applied to prove the concept is given (Sec.~V). After showing the implementation on a prismatic and an articulated monopod hopper with different actuation (Sec.~VI and~VII), we conclude the established framework (Sec.~VIII).}

\section{Context and Problem Statement}
\label{sec:ProblemStatement}
%
In the following, we present a formal description of the gait optimization problem \revrm{that we are addressing in this paper.
This description includes}\rev{including} a definition of the class of hybrid dynamical systems, and our notion of periodicity.

\subsection{Hybrid Dynamics of Legged Systems}
We model a legged system as a mechanical rigid multi-body system using a floating-base description.
This system is subject to unilateral contact without sliding, similarly to the systems presented in \cite{raff2022generating}. The vector $\mathbf{q} \in \mathcal{Q} \subset \mathbb{R}^{n_\mathrm{q}}$ describes the configuration of the system with $n_\mathrm{q}$ degrees of freedom within the configuration space $\mathcal{Q}$. The state of the system is given with the vector $\mathbf{x} = (\mathbf{q}, \dot{\mathbf{q}}) \in T\mathcal{Q} \subset \mathbb{R}^{2n_\mathrm{q}}$, where $T\mathcal{Q}$ is the tangent bundle of $\mathcal{Q}$. The system is operated with an input $\mathbf{u} \in \mathbb{R}^{n_\mathrm{u}}$, where $n_\mathrm{u}$ is the number of actuated joints. We obtain the differential-algebraic equation of the form
\begin{subequations}\label{eq:continuous}
    \begin{align}
        \mathbf{M}(\mathbf{q}) \ddot{\mathbf{q}} &= \mathbf{n}(\mathbf{q},\dot{\mathbf{q}}) + \mathbf{J}(\mathbf{q})\T \mathbf{u} + \mathbf{W}_i(\mathbf{q}) \boldsymbol{\uplambda}_i, 
        \label{eq:EoM}
        \\
        \mathbf{g}_i(\mathbf{q}) &= \mathbf{0},
        \label{eq:contact}
    \end{align}
\end{subequations}
with mass matrix $\mathbf{M}$ and generalized forces $\mathbf{n}$. The input vector is mapped by the \rev{selection matrix} $\mathbf{J}\T$. Active contacts are expressed in \eqref{eq:contact} as holonomic bilateral constraints $\mathbf{g}_i$, where $\mathbf{W}_i(\mathbf{q})\T := \partial \mathbf{g}_i / \partial \mathbf{q}$ is the corresponding constraint Jacobian with constraint forces $\boldsymbol{\uplambda}_i$. For subsequent use, we rewrite the equations of motion \eqref{eq:EoM} as first-order differential equation 
\begin{equation}
    \dot{\x} = \mathbf{f}_i(\x,\bu).
    \label{eq:EoMGeneral}
\end{equation}
The set of active constraints in \eqref{eq:contact} depends on the current \textit{phase} $i \in \mathcal{I}$, where $\mathcal{I}$ is a finite set of contact configurations. In case of touch-down or lift-off, phase transitions occur. These are triggered by event functions $e_i^j(\mathbf{x}) = 0$, where $e_i^j \in C^2$. We project the state from phase $i$ to $j$ with a discrete map
\begin{equation}\label{eq:discrete}
    \boldsymbol{\Delta}_i^j(\mathbf{x}) = \begin{bmatrix}
        \mathbf{q} \\
        \big(\mathbf{I} - \mathbf{M}^{-1} \mathbf{W}_j(\mathbf{W}_j\T \mathbf{M}^{-1} \mathbf{W}_j )^{-1} \mathbf{W}_j\T \big) \, \dot{\mathbf{q}}
    \end{bmatrix}.
\end{equation}
Thus, we can map the states right before and after the event $e_i^j$ with $\mathbf{x}^+ = \boldsymbol{\Delta}_i^j(\mathbf{x}^-)$ \cite{brogliato2016nonsmooth}. 
We assume the existence of a hybrid trajectory $\mathbf{x}(t)$ passing through a predefined order of phases consisting of continuous~\eqref{eq:continuous} and discrete~\eqref{eq:discrete} dynamics.
Further, we assume that the sequence of phases remains fixed under local changes within $\mathbf{x}(t)$.
With this assumption, we can separate the hybrid trajectory and, particularly, its input space into $m \in \mathbb{N}$ phases.\par
For each phase $i$, we parameterize the input $\bu_i(t,\x,\bxi_i)$, with $\bu_i:\mathbb{R}\times T\mathcal{Q}\times \mathcal{U}_{\xi,i}\to \mathbb{R}^{n_\mathrm{u}}$, where $\bxi_i \in \mathcal{U}_{\xi,i}$ is a suitable parameterization of a time- and/or state-based control law. Importantly, we assume a parameterization that fulfills $\bu_i(t,\x,\mathbf{0}) \equiv \mathbf{0}$ for all $i \in \mathcal{I} $, $t \in \mathbb{R}$, and $\x\in T\mathcal{Q}$. Note that the dependencies of $\bu_i$ indicate the possibility of a feedforward (dependency on $t$) as well as a feedback (dependency on $\x$) control input. The shape of the hybrid trajectory $\x(t)$ is only affected by the initial state $\x_0$ of the system and the input parameters $\bxi \in \mathcal{U}_\xi$, with $\mathcal{U}_{\xi}=\bigcup_{i=1}^{m} \mathcal{U}_{\xi,i}$, collecting $\bxi_i$ over all $m$ phases. Henceforth, we indicate the influence on the trajectory with the dependencies $\x(t;\x_0,\bxi)$.

\subsection{Gaits and the Implied Periodicity}
We consider gaits as periodic motions with period time $T > 0$. In these trajectories, one full period is called a \textit{stride}. To account for aperiodic states (e.g., horizontal position in forward motion), we introduce a periodicity matrix~$\mathbf{P} \in \mathbb{R}^{2n_\mathrm{q} \times 2n_\mathrm{q}}$ to select periodic states. It also facilitates symmetric gait search by permuting state order within the periodicity condition. With this matrix, we define periodicity as $\mathbf{P}\,\x(T) =\x_0$.
Additionally, we define the two $C^2$ functions $\revMath{\rho} : T\mathcal{Q} \rightarrow \mathbb{R}$ and $O :T\mathcal{Q} \times \mathbb{R} \times \mathbb{R}\rightarrow \mathbb{R}$. The first introduces an anchor constraint $\revMath{\rho}(\x(T)) = 0$, which describes the requirement for the end of a stride, for example, touch-down of the foot. With the second, we can use the implicit function $O(\x(t;\x_0,\bxi),T,\bar{O}) = 0$ to limit the motion to an operating point $\bar{O}$. For example, a desired average speed for a gait.
All constraints can be put together to define a \textit{gait} by the roots of the function $\mathbf{h}(\x(t;\x_0,\bxi),T,\bar{O}) = \mathbf{0}$:
\begin{equation}
    \mathbf{h}(\cdot) = \begin{bmatrix}
        \mathbf{P}\,\x(T;\x_0,\bxi)-\x_0 \\
        \revMath{\rho}\big(\x(T;\x_0,\bxi)\big) \\
        O\big(\x(t;\x_0,\bxi),T,\bar{O}\big)
    \end{bmatrix}
    \label{eq:ConstraintGeneral}
\end{equation}
where $\x_0 = \x(t=0)$.

\subsection{Optimally Actuated Gaits}
\label{sec:optimality}
The overall objective of this paper is to find optimal control inputs for such periodic gaits.
To this end, we define a cost function $c : T \mathcal{Q} \times \mathcal{U}_\xi \times \mathbb{R} \rightarrow \mathbb{R}$ that we seek to minimize.
As we are interested primarily in energetically economic motions, we can make the following assumptions about this cost function: 
\begin{enumerate}
    \item it is at least twice differentiable in all variables, i.e., $c \in C^2$,
    \item it is of zero value if $\mathbf{u} \equiv \mathbf{0}$, i.e., $c(\x_0,\bxi = \mathbf{0},T) = 0$, and
    \item locally positive-definite in $\frac{\partial^2 c}{\partial \bxi^2} (\x(\cdot)) |_{\bxi = \mathbf{0}}$.
\end{enumerate}
A constrained optimization problem is defined by coupling the energetic cost function $c$ with the gait constraints (\ref{eq:ConstraintGeneral}):
\begin{equation}
    \label{eq:COP}
    \begin{split}
        \min_{\mathbf{a}} \quad &c(\mathbf{a}) \\
        \text{subject to} \quad &\mathbf{h}(\mathbf{a}, \Bar{O}) = \mathbf{0},
    \end{split}
\end{equation}
with the decision variable $\ba\T = [\x_0\T,\bxi\T,T]$. 

\section{Passive and Quasi-Passive Systems}
\label{sec:QuasiPassiveSystems}

\subsection{Passive Gaits}
We refer to the subset of solutions of \eqref{eq:ConstraintGeneral} with no actuation (i.e.,  $(\x(t;\x_0,\mathbf{0}),T,\bar{O}) \in \mathbf{h}^{-1}(\mathbf{0})$) as the set of \textit{passive} gaits.
Apart from isolated solutions in specifically design mechanisms\footnote{Which exhibit, for example, perfect ground speed matching to eliminate collision losses \cite{chatterjee2002persistent}.}, the existence of passive motions on level ground generally requires that the underlying system is energetically conservative, as in \cite{raff2022connecting}.
This can be formalized via the total energy in the system $E: T\mathcal{Q} \rightarrow \mathbb{R}$, which must be constant for energetically conservative systems.
That is, it holds:
\begin{equation}
    E \big( \x(t;\x_0,\bxi=\mathbf{0}) \big) = \mathrm{const.}, \quad \forall t,\x_0.
\end{equation}
It has been demonstrated that such passivity can be a very valuable tool in gait generation.
In particular, one can use numerical continuation methods to systematically find one-dimensional manifolds of passive solutions to \eqref{eq:ConstraintGeneral}, which form families of connected periodic motions \cite{raff2022connecting}. 
If parameterized by the operating point~$\bar{O}$, these families can form a useful library of different gaits.

\subsection{Quasi-Passivity}
\label{sec:QuasiPassivity}
Naturally, real robotic systems are dissipative and undergo energy losses during their motion:
\begin{equation}
    \dot{E}\big(\x(t;\x_0,\bxi=\mathbf{0})\big) \leq 0, \quad \forall t,\x_0
\end{equation}
Without compensation, this loss of energy in the system prevents periodicity for the states $\mathbf{P} \x$, and thus makes finding gaits impossible without using actuation $\bxi \neq \mathbf{0}$.

In order to employ the above mentioned ideas to dissipative systems with friction and impacts, we will compensate the dissipation via a \emph{virtual} energy injecting input $\mathbf{f}_\mathrm{E}(\x,\gamma) \in \mathbb{R}^{2 n_\mathrm{q}}$, with:
\begin{subequations}
\begin{equation}
    \dot{\mathcal{E}}\big(\mathbf{f}_\mathrm{E}(\x,\gamma)\big) \overset{!}{\geq} 0, \quad \forall \x
    \label{eq:VirtualInputCond}
\end{equation}
where $\dot{\mathcal{E}} : \mathbb{R}^{2n_\mathrm{q}} \rightarrow \mathbb{R}$ represents the power that is inserted into the system.
One should note that such an input must not necessarily be realizable on a physical robot.
This virtual input term is inserted additively into the equations of motion, extending~\eqref{eq:EoMGeneral} to:
\begin{equation}
    \dot{\x} = \mathbf{f}_i(\x,\bu_i) +  \mathbf{f}_\mathrm{E}(\x,\gamma).
\end{equation} 
\end{subequations}

For $\bu_i\equiv \mathbf{0}$, we refer to such systems as being \emph{quasi-passive}.
The central idea of this work is to use quasi-passivity in non-conservative systems to find families of periodic motions that are then connected to active optimal solutions. 
We scale the injected amount of energy with one single \rev{input} parameter $\{\gamma \in \mathbb{R} | \gamma > 0 \}$ that we carefully tune, such that a desired gait can be executed as a periodic orbit in a dissipative system despite the absence of actuation $\boldsymbol{\xi} = \mathbf{0}$.
For simplicity, we assume that this energy injection is independent of the current contact configuration in phase~$i$ and thus only depends on the state $\x$.

There exists a number of different ways to achieve such energy injection.
The most prominent example would be to put the legged system onto downward sloping ground, where the parameter $\gamma$ would represent the slope angle.
As long as the center of gravity is moving monotonically forward, the resulting increase in the gravitational potential injects energy, which can compensate friction and impact losses.
Such locomotion on slopes has been studied extensively for so-called passive dynamic walkers \cite{mcgeer1990passive} and has been the base for a systematic generation of families of gaits in \cite{rosa2023approach}.

Here, we seek to generalize this idea by considering different ways to inject energy.
For example, one could use the idea of~\cite{sepulchre1997localized}, by defining:
\begin{subequations}
\begin{equation}
    \mathbf{f}_\mathrm{E}(\x,\gamma) := \gamma \cdot \nabla E(\x)=\gamma \cdot \begin{bmatrix}
        \frac{\partial E }{\partial \q}\T\\[1mm] \mathbf{M}(\q) \, \dq
    \end{bmatrix}, 
    \label{eq:EnrgyInjSepulchre}
\end{equation}
where \eqref{eq:VirtualInputCond} is fulfilled for any nonnegative~$\gamma$:
\begin{equation}
    \dot{\mathcal{E}}= \nabla E(\x)\T \mathbf{f}_\mathrm{E}(\x,\gamma) = \gamma \cdot  \Vert \nabla E(\x) \Vert_2^2 \geq 0.
\end{equation}
\end{subequations}

Alternatively, we could simply use uniform negative damping on all joints as the virtual energy input, with $\gamma$ being the negative damping coefficient:
\begin{equation}
    \mathbf{f}_\mathrm{E}(\x,\gamma) := \gamma \cdot\begin{bmatrix}
        \mathbf{0}\\ \mathbf{M}(\q)^{-1} \, \dq
    \end{bmatrix},\quad \dot{\mathcal{E}}=\gamma \cdot  \Vert \dq\Vert_2^2 \geq 0.
    \label{eq:EnrgyInjUni}
\end{equation}
Note that in this formulation the damping force $\gamma\dq$ is multiplied by $\mathbf{M}(\q)^{-1}$ to yield acceleration values.

Finally, we can use mass-proportional damping to maintain the decoupling of nonlinear normal modes of dissipative systems as in \cite{krack2015nonlinear}.
An injection of this form does not have any local influence on the natural frequencies or the mode shapes of the system:
\begin{equation}
    \mathbf{f}_\mathrm{E}(\x,\gamma) := \gamma \cdot\begin{bmatrix}
        \mathbf{0}\\ \dq
    \end{bmatrix},\quad \dot{\mathcal{E}}=\gamma \cdot  \underbrace{\dq\T \mathbf{M}(\q)\dq}_{=2 E_\mathrm{kin}(\x)} \geq 0.
    \label{eq:EnrgyInjPMC}
\end{equation}
Again, the mass-scaled damping force $\gamma\mathbf{M}(\q)\dq$ has been multiplied by $\mathbf{M}(\q)^{-1}$ to yield acceleration values.

With this extension, we are now able to find periodic motions for dissipative systems in the total absence of actual actuator inputs ($\bu_i\equiv \mathbf{0}\Leftrightarrow\bxi_i = \mathbf{0}$).
To this end, we need to determine only one scalar parameter $\gamma$, which tunes the virtual energy injection and compensates for all losses. 

\section{Optimal Actuation via Homotopic Continuation}
\label{sec:HomotopicContinuation}

As mentioned above, quasi-passive gaits are not necessarily realizable on physical systems, which may not have actuation on all joints.
We thus need to establish a connection between the quasi-passive gaits and optimally actuated gaits.
Therefore, we introduce the concept of an \textit{input homotopy}, which describes a mapping from a model including virtual energy injection to a model with only motor actuation as input. This homotopy is employed to map quasi-passive gaits to actuated gaits of a physical model using a homotopy parameter $\varepsilon \in [0,1]$. We scale the virtual input as follows:
\begin{equation}
    \dot{\x} = \mathbf{f}_i(\x,\bu_i) + (1-\varepsilon)\cdot\mathbf{f}_\mathrm{E}(\x,\gamma)
    \label{eq:homotopy}
\end{equation}
Hence, with $\varepsilon = 0$ the system is fully supported by virtual energy injection. A homotopic continuation approach is started from a solution $(\x(t;\x_0,\bxi),T,\bar{O})$ of the gait constraints~\eqref{eq:ConstraintGeneral} with $\varepsilon = 0$, for which no actuation profile has to be determined, as $\bxi = \mathbf{0}$ by definition.
This can be done with less effort using common root-search algorithms, similar to \cite{raff2022connecting,rosa2021topological}. Consequently, for continuation, there is no need for initial guesses from the user.
Introducing the parameter~$\varepsilon$, we can redefine the optimization problem~\eqref{eq:COP} as a parameterized optimization problem:
\begin{equation}
    \label{eq:POP}
    \begin{split}
        \min_{\mathbf{a}} \quad &c(\mathbf{a},\varepsilon) \\
        \text{subject to} \quad &\mathbf{h}(\mathbf{a}, \Bar{O},\varepsilon) = \mathbf{0}.
    \end{split}
\end{equation}
We can observe that $\varepsilon = 0$ defines a family of optimal solutions, for which we have already found one through root-search. This quasi-passive solution is our starting point for a problem that can be solved by continuation. The resulting solution at $\varepsilon = 1$ contains the trajectory $\x(t;\x_0,\bxi)$ of a fully actuated gait. Also, the control actions are adapted to match this gait $(\bxi \neq \mathbf{0})$. Combining the method of virtual energy injection and the continuation via input homotopy allows us to find optimally actuated gaits for dissipative systems, where the losses are counteracted through the actuator forces.


\section{Framework and Implementation}
\label{sec:Implementation}
The overall concept of our approach to derive active gaits for dissipative systems is summarized in Figure~\ref{fig:Concept}.
We combine the different methods into a three-stage approach:
\begin{enumerate}
    \item We begin by injecting virtual energy into the model of a dissipative system. We use energy injection of the form \eqref{eq:EnrgyInjPMC}, based on the mass-proportional damping approach from~\cite{krack2015nonlinear}. This energy injection directly affects the nonlinear equations of motion in each dimension, regardless of whether physical actuation of each state would be possible. This makes it possible to support the movement of a system with only one single parameter. We can interpret the quasi-passive model (Sec.~\ref{sec:QuasiPassivity}) as \eqref{eq:homotopy} with $\varepsilon=0$.
    \item We then apply a root-finding algorithm to find quasi-passive gaits. To this end, we solve the \revrm{nonlinear program (}NLP\revrm{)} defined by \eqref{eq:ConstraintGeneral} with the additional constraint $\bxi = \mathbf{0}$ while simultaneously determining~$\gamma$. Note that whole families of quasi-passive gaits \revrm{with selected properties }can be found using numerical continuation algorithms~\cite{raff2022connecting, rosa2023approach}.
    \item Homotopic continuation is applied to obtain an optimally actuated gait. The given quasi-passive solution is continued until the virtual energy injection is fully deactivated, i.e., $\varepsilon=1$ (Sec.~\ref{sec:HomotopicContinuation}). 
\end{enumerate}

We implement this framework with MATLAB and CasADi~\cite{Andersson2019}, a tool for nonlinear optimization and algorithmic differentiation. In the following, the applied parameterization, root-search, and continuation algorithms are specified.

\subsection{Direct Collocation}

For the root-search, we choose a direct collocation algorithm, which allows us to discretize the system dynamics and include it as collocation constraints in the NLP. The system dynamics is thus approximated as polynomial splines. The discretization offers a way to parameterize the control inputs $\bu_i(t,\bxi)$ as piecewise constant feedforward control. We specify the parameterization vector $\bxi_i \in \mathbb{R}^{(n_\mathrm{u} \cdot N)} \subset \mathcal{U}_{\xi,i}$, where $N$ is the number of discretized time intervals over phase~$i$. We define $\bxi_i$ to contain the constant values of $\bu_i(t)$ over every time interval $\bxi_i\T = [\bu\T_{i,1}, \bu\T_{i,2}, \dots , \bu\T_{i,N}]$. We find the root of \eqref{eq:ConstraintGeneral} and the collocation constraints with the NLP solver IPOPT and obtain a quasi-passive gait with the adapted virtual energy parameter~$\gamma$ and actuation $\bxi = \mathbf{0}$.

\subsection{Continuation on First-Order Optimality}

\begin{figure}
    \centering
    \resizebox{0.75\linewidth}{!}{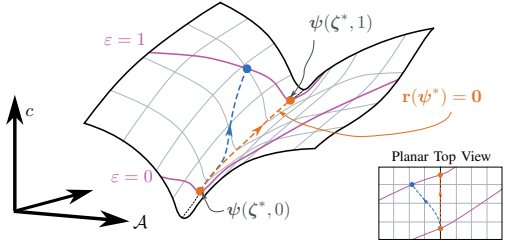}
    \caption{Continuation on the first-order optimality. The coordinate system indicates the cost $c$ in the vertical and the search space $\mathcal{A}$ in the remaining dimensions. The continuation starts at the regular point $\boldsymbol{\psi}(\boldsymbol{\zeta}^*,\varepsilon = 0)$ -- a solution, which we call \textit{quasi-passive}. From there, we move along a one-dimensional manifold (orange), where the first-order optimality condition $\mathbf{r}(\boldsymbol{\psi}^*) = \mathbf{0}$ is fulfilled. Note that we advance in the direction of increasing~$\varepsilon$. The regular point $\boldsymbol{\psi}(\boldsymbol{\zeta}^*,\varepsilon = 1)$ marks the \revrm{obtained }result with a locally optimal cost. The blue curve shows a \revrm{possible }path towards $\varepsilon = 1$, which is not optimal.\revrm{ In the lower right corner, a top view of the surface is depicted.}}
    \label{fig:Continuation}
\end{figure}

We want to exploit the introduced input homotopy to obtain optimally actuated gaits. Thus, we apply a pseudo-arclength continuation \cite{allgower2003introduction} on the first-order optimality of the parameterized optimization problem. To define this optimality, the Lagrangian for \eqref{eq:POP} is specified as
\begin{equation}
    \label{eq:Lagrangian}
    L(\bzeta, \bar{O},\varepsilon) = c(\ba,\varepsilon) + \blambda\T \, \mathbf{h}(\ba, \bar{O},\varepsilon),
\end{equation}
and the first-order optimality condition as
\begin{equation}
    \label{eq:FirstOrdCond}
    \frac{\partial L}{\partial \ba} (\bzeta, \bar{O},\varepsilon) = \frac{\partial c}{\partial \ba}(\ba,\varepsilon) + \blambda\T \frac{\partial \mathbf{h}}{\partial \ba}(\ba,\bar{O},\varepsilon) = \mathbf{0},
\end{equation}
where the vector $\bzeta\T = [\ba\T, \blambda\T]$ describes an extended decision variable. The vector $\blambda \in \mathbb{R}^{n_h}$ is the Lagrange multiplier. We can deduce the uniqueness of $\blambda$ from vector~$\ba$ being a local extremum point of $c$, when it is subject to the constraints $\mathbf{h}(\cdot) = \mathbf{0}$, and from $\partial \mathbf{h}/\partial \ba$ having full rank \cite{luenberger1984linear}.\par

Let $\br : \mathbb{R}^{n_\zeta+1} \to \mathbb{R}^{n_\zeta}$ be an implicit function, which we call \textit{homotopy map}. A point $\bpsi(\bzeta^*,\varepsilon^*)\in\mathbb{R}^{n_\zeta+1}$ is called a regular point of $\br$ if $(\frac{\partial\br}{\partial\bpsi})|_{\bpsi = \bpsi^*}$ has maximal rank $n_\zeta$. A quasi-passive solution $(\bzeta^*,0)$ is a regular point of $\br$. Assume that both the quasi-passive and the actuated model are connected through a one-dimensional manifold --~the solution curve $\bpsi$~--, on which the homotopy map $\br(\bpsi^*)=\mathbf{0}$ is always fulfilled as a condition. The curve $\bpsi$ can be numerically traced with predictor-corrector methods~\cite{allgower2003introduction}. We make a predictor step in the tangent space of the curve and correct this step so that the condition of the homotopy map is again fulfilled. We start from $\varepsilon = 0$ and trace the condition in $\br$ into a unique direction, where $\varepsilon$ is increasing. 

For the predictor the tangent vector $\mathbf{p}$ at a regular point $(\boldsymbol{\zeta}^*, \varepsilon^*)$ is uniquely defined by
\begin{subequations}
    \label{eq:TangentVector}
    \begin{align}
        \mathbf{R}(\boldsymbol{\zeta}^*, \varepsilon^*) \cdot \mathbf{p} = \mathbf{0}, \\
        \Vert \mathbf{p} \Vert_2 = 1, \\
        \det \Biggl( 
        \begin{bmatrix}
            \mathbf{R}(\boldsymbol{\zeta}^*, \varepsilon^*) \\
            \mathbf{p}\T
        \end{bmatrix}    
        \Biggr) > 0.
    \end{align}
\end{subequations}
The matrix $\mathbf{R}$ is the partial derivative of the homotopy map~$\mathbf{r}$ with respect to $\boldsymbol{\zeta}$ and $\varepsilon$:
\begin{equation}
    \label{eq:R}
    \mathbf{R}(\boldsymbol{\zeta}, \varepsilon) = \frac{\partial \mathbf{r} \, (\boldsymbol{\zeta}, \varepsilon)}{\partial (\boldsymbol{\zeta}, \varepsilon)}.
\end{equation}
\rev{While we apply automatic differentiation \cite{Andersson2019} to obtain second-order derivatives for $\mathbf{R}$, an analytical description as in \cite{singh2023analytical} might be possible, as well.}

To perform continuation on the first-order optimality the condition \eqref{eq:FirstOrdCond} is included in the homotopy map:
\begin{equation}
\label{eq:HomotopyMap}
    \br(\bzeta, \varepsilon) := \frac{\partial L_{\bar{O}}}{\partial \bzeta}(\bzeta, \varepsilon) = 
    \begin{bmatrix}
        (\frac{\partial c}{\partial \ba})\T + (\frac{\partial \mathbf{h}}{\partial \ba})\T \blambda \\
        \mathbf{h}(\cdot)
    \end{bmatrix}.
\end{equation}
The subscript notation $L_{\bar{O}}(\boldsymbol{\zeta},\varepsilon)$ indicates a fixed operating point $\bar{O}$ in \eqref{eq:Lagrangian}. We start the predictor-corrector method from a quasi-passive gait, which is a regular point of $\br$ by definition, as $\bxi = \mathbf{0}$, and we defined $c(\x_0,\bxi = \mathbf{0},T) = 0$ in Sec.~\ref{sec:optimality}. We deform the solution in the direction of positive~$\varepsilon$ while remaining at local minima.
At $\bpsi(\bzeta^*,1)$ an optimally actuated solution with $\bxi^* \neq \mathbf{0}$ is computed without any virtual support. To prove that the regular point is also a strict minimizer, we need the second-order optimality condition as a sufficient condition:
\begin{equation}
    \label{eq:SecondOrdOpt}
    \mathbf{y}\T \, \Biggl( \frac{\partial^2 c}{\partial \ba^2} (\ba,\varepsilon) + \blambda\T \frac{\partial^2 \mathbf{h}}{\partial \ba^2} (\ba, \bar{O},\varepsilon)\Biggr) \, \mathbf{y} > 0,
\end{equation}
$\forall\mathbf{y} \in Y$, where we define the tangent space in $\ba^*$ as $Y = \{\mathbf{y}: (\partial \mathbf{h} / \partial \ba) |_{\ba = \ba^*} \mathbf{y} = \mathbf{0}\}$ and $\ba^*$ is part of the regular point $\bzeta^*$ with fixed $\bar{O}$. For conditions \eqref{eq:FirstOrdCond} and \eqref{eq:SecondOrdOpt} it is essential that the equality constraints $\mathbf{h}$ in \eqref{eq:COP} are regular to find strict local minima \cite{luenberger1984linear}.\par
Figure \ref{fig:Continuation} depicts a schematic of the continuation approach. 
An algorithm of the complete procedure of continuation on the first-order optimality can be found in~\cite{raff2022generating}. \rev{The presented method is applicable to a variety of legged systems. Its complexity scales quadratically with $n_\mathrm{q}$ and the complete number of grid points $m\cdot N$.} In the following sections, we present the results of the implemented approaches on two compliant and dissipative systems\footnote{\rev{A detailed mathematical description of the examples and the implementation of the approach is given in \url{arxiv.org/abs/xxxx.xxxxx}.}}.

\begin{table}[]
    \centering
    \begin{tabular}{c l c l}
        \hline
         $\revMath{\x}$ & \rev{system states} & $\revMath{\gamma}$ & \rev{virtual energy parameter} \\
         $\revMath{\bxi}$ & \rev{input parameters} & $\revMath{\mathbf{f}_E}$ & \rev{virtual input term} \\
         $\revMath{\ba}$ & \rev{decision variable} & $\revMath{\varepsilon}$ & \rev{homotopy parameter} \\
         $\revMath{\bzeta}$ & \rev{extended decision var.} & $\revMath{\br}$ & \rev{homotopy map} \\
         $\revMath{\mathbf{h}}$ & \rev{constraint function} & $\revMath{\mathbf{p}}$ & \rev{tangent vector} \\
         $\revMath{c}$ & \rev{cost function} & $\revMath{\bpsi}$ & \rev{solution curve} \\
         \hline
    \end{tabular}
    \rev{\caption{Notation of Selected Quantities}}
\end{table}

\section{Example: Prismatic Monopod}
\label{sec:Prism}

\begin{figure}
    \centering
    \resizebox{0.7\linewidth}{!}{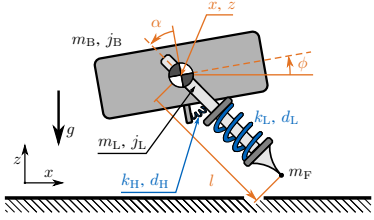}
    \caption{The prismatic monopod model contains a main body with mass $m_\mathrm{B}$ and inertia $j_\mathrm{B}$, a leg with mass $m_\mathrm{L}$ and inertia $j_\mathrm{L}$ and a foot as point mass $m_\mathrm{F}$. There is a radial spring with stiffness $k_\mathrm{H}$ and damping $d_\mathrm{H}$ at the hip joint, as well as a linear spring with stiffness $k_\mathrm{L}$ and damping $d_\mathrm{L}$ between the leg and foot. The configuration of the system is described by $\mathbf{q} = [x\,z\,\phi\,\alpha\,l]\T$.}
    \label{fig:PrismaticMonopod}
\end{figure}

We apply our methodology to a prismatic monopod model, similar to the one proposed in \cite{remy2011optimal}.
The model is composed of a main body with mass $m_\mathrm{B}$ and inertia $j_\mathrm{B}$. The variables~$x$ and $z$ describe the horizontal and vertical positions of the CoM, respectively. The main body's pitch angle is denoted by $\phi$. The leg is mounted on a rotational joint positioned on the main body's CoM. The joint angle between the main body and leg is $\alpha$. The two parts are connected with a torsional spring with stiffness~$k_\mathrm{H}$ and damping~$d_\mathrm{H}$. The spring is uncompressed for $\alpha = 0$. The leg has mass~$m_\mathrm{L}$ and inertia~$j_\mathrm{L}$ and is connected to a foot with a translational linear spring. The foot is at radial position $l$ relative to the CoM of the main body. The generalized coordinates are thus $\mathbf{q} = [x, z, \phi,\alpha,l]\T$, with $n_\mathrm{q} = 5$. The translational spring is uncompressed at $l = l_0$ and has stiffness~$k_\mathrm{L}$ and damping~$d_\mathrm{L}$. Parallel to both springs, there are actuators that produce the torques $\mathbf{u} = [u_\alpha,u_\mathrm{l}]\T$. The foot is modeled as a point mass with mass $m_\mathrm{F}$. It is important to mention that $m_\mathrm{F} > 0$ since a non-zero foot mass almost entirely prevents energetic conservativeness even with $d_\mathrm{H} , d_\mathrm{L} = 0$. Figure \ref{fig:PrismaticMonopod} shows the system and its components. All values are normalized with respect to leg length $l_0$, total mass $m_0$, and gravity $g$.\par

\subsection{Quasi-Passive Gait}
\label{sec:ResultsPrismQuasi}
We utilize the direct collocation approach with $N=10$ per phase to find quasi-passive gaits. Note that damping and collision losses are included in the motion of the system. Virtual energy injection is used to achieve a quasi-passive motion, i.e., we reach the same total energy at the end of a stride as in the beginning. Thus, the continuation parameter is set to $\varepsilon = 0$, and the parameter, which controls the amount of virtual energy input, is initialized to $\gamma_{\text{init}} = 0.01$. Over the root-search, this parameter changes to a value $\gamma > 0$, such that the losses through damping and impacts are compensated.\par
\begin{figure}
\begin{minipage}{0.05\linewidth}
    \centering
    \rotatebox{90}{\footnotesize Continuous States}
    \vspace{0.4cm}
\end{minipage}
\begin{minipage}{0.95\linewidth}
\pgfplotsset{footnotesize}
    \begin{tikzpicture}[baseline]
        \begin{axis}[
            title={Prismatic Monopod: Quasi-Passive Gait},
            xtick={0,.5,1,1.5},
            ytick={-.2,0,.2,.4,.6,.8,1,1.2},
            xmin=0,
            xmax=1.9,
            ymin=-0.3,
            ymax=1.2,
            legend pos=outer north east,
            grid = major,
            legend entries={$x$,$z$,$\phi$,$\alpha$,$l$},
            width=7cm,
            height=3.5cm
        ]
    
            \addplot[tumBlue,line width=1.2pt] table [
            col sep=comma,
            mark=none
            ] {Data/Figure02_1/Figure02_1_x.csv};
        
            \addplot [tumOrange,line width=1.2pt] table [
            col sep=comma, 
            mark=none
            ] {Data/Figure02_1/Figure02_1_z.csv};
    
            \addplot [tumYellow,line width=1.2pt] table [
            col sep=comma, 
            mark=none
            ] {Data/Figure02_1/Figure02_1_phi.csv};
    
            \addplot [tumPink,line width=1.2pt] table [
            col sep=comma, 
            mark=none
            ] {Data/Figure02_1/Figure02_1_gamma.csv};
    
            \addplot [tumGreen,line width=1.2pt] table [
            col sep=comma, 
            mark=none
            ] {Data/Figure02_1/Figure02_1_l.csv};

            \path[name path=axis] (0,-1) -- (0,2);
            \path[name path=f] (0.77,-1) -- (0.77,2);
            \addplot[
                color=gray,
                fill opacity=0.2
            ] fill between[
                of=axis and f
            ];

        \end{axis}
    \end{tikzpicture}
\hspace{0.15cm}
\pgfplotsset{footnotesize}
    \begin{tikzpicture}[baseline]
        \begin{axis}[
            xlabel={Time $[\sqrt{l_0 / g}]$},
            xtick={0,.5,1,1.5},
            ytick={-.6,-.4,-.2,0,.2,.4,.6,.8,1},
            xmin=0,
            xmax=1.9,
            ymin=-0.7,
            ymax=1.1,
            legend pos=outer north east,
            grid = major,
            legend entries={$\dot{x}$,$\Dot{z}$,$\Dot{\phi}$,$\Dot{\alpha}$,$\Dot{l}$},
            width=7cm,
            height=3.5cm
        ]
    
            \addplot[tumBlue,line width=1.2pt] table [
            col sep=comma,
            mark=none
            ] {Data/Figure02_2/Figure02_2_dx.csv};
        
            \addplot [tumOrange,line width=1.2pt] table [
            col sep=comma, 
            mark=none
            ] {Data/Figure02_2/Figure02_2_dz.csv};
    
            \addplot [tumYellow,line width=1.2pt] table [
            col sep=comma, 
            mark=none
            ] {Data/Figure02_2/Figure02_2_dphi.csv};
    
            \addplot [tumPink,line width=1.2pt] table [
            col sep=comma, 
            mark=none
            ] {Data/Figure02_2/Figure02_2_dgamma.csv};
    
            \addplot [tumGreen,line width=1.2pt] table [
            col sep=comma, 
            mark=none
            ] {Data/Figure02_2/Figure02_2_dl.csv};

            \path[name path=axis] (0,-1) -- (0,2);
            \path[name path=f] (0.77,-1) -- (0.77,2);
            \addplot[
                color=gray,
                fill opacity=0.2
            ] fill between[
                of=axis and f
            ];
        \end{axis}
    \end{tikzpicture}
\end{minipage}
\caption{The continuous states of the hopping forward gait of the prismatic monopod model, including virtual energy injection. The operating point constraint is set to the average speed $\Bar{\Dot{x}} = 0.3 \sqrt{l_0g}$. The gray area marks the stance phase. The main body swings over a stride and compensates the back and forward motion of the leg.}
\label{fig:PrismForward}
\end{figure}
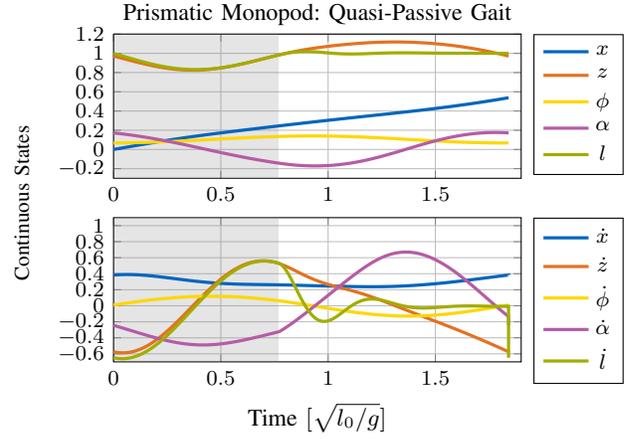
\begin{figure}
    \centering
    \includegraphics[width=0.9\linewidth]{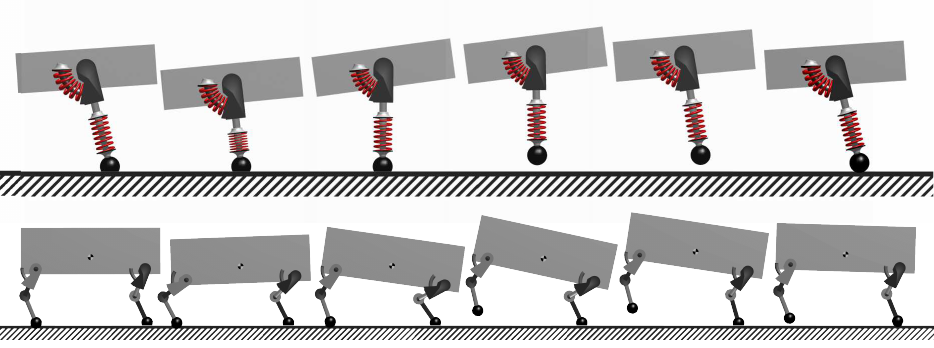}
    \caption{Snapshots of the quasi-passive hopping forward gait of the prismatic monopod (top, Sec. \ref{sec:Prism}) and \revrm{the segmented monopod (bottom, Sec. ?) }\rev{of the bounding gait of the sagittal quadruped (bottom, Sec. \ref{sec:Quadru}).}}
    \label{fig:SnapshotPrism}
\end{figure}

We start the root-search for a hopping forward gait with average speed $\bar{\dot{x}} = 0.3 \sqrt{l_0g}$. \rev{As an initial guess, we propose a non-periodic motion, which fulfills the correct sequence of events (\textit{lift-off} $\rightarrow$ \textit{touch-down}).} Figure~\ref{fig:PrismForward} depicts the trajectories of the individual states. Figure~\ref{fig:SnapshotPrism} shows snapshots of the resulting motion. 
We can observe a forward motion of the monopod. The main body leans to the left and swings over the stride to compensate for the rotation of the leg, which swings back in stance and forth in flight. The main body moves horizontally forward with the given average speed. At the end of the stride, all states reach their initial value again, thus, the orbit is periodic. Note that it is possible to utilize continuation to identify families of gaits on different operating points, from where the homotopic continuation can be initialized.

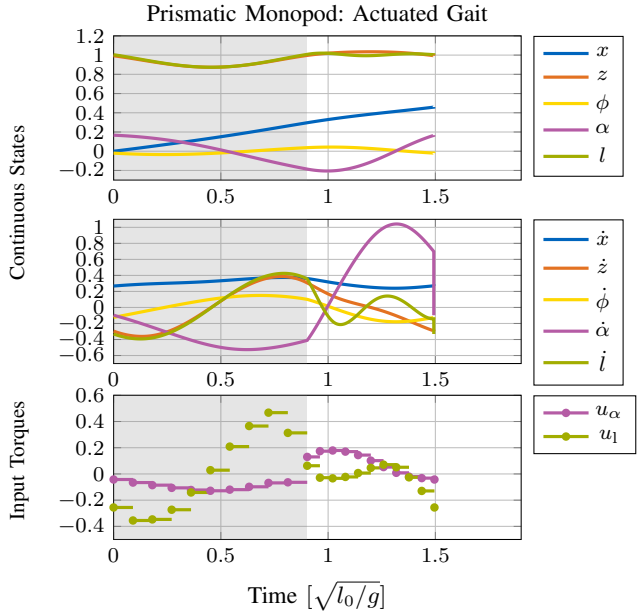
\begin{figure}
\begin{minipage}{0.05\linewidth}
    \rotatebox{90}{\footnotesize Input Torques \hspace{1.4cm} Continuous States \hspace{0.4cm}}
\end{minipage}
\begin{minipage}{0.95\linewidth}
    \pgfplotsset{footnotesize}
    \begin{tikzpicture}
        \begin{axis}[
            name=main plot,
            title={Prismatic Monopod: Actuated Gait},
            xtick={0,.5,1,1.5},
            ytick={-.2,0,.2,.4,.6,.8,1,1.2},
            xmin=0,
            xmax=1.9,
            ymin=-0.3,
            ymax=1.2,
            legend pos=outer north east,
            grid = major,
            legend entries={$x$,$z$,$\phi$,$\alpha$,$l$},
            width=7cm,
            height=3.5cm
        ]

            \addplot[tumBlue, line width=1.2pt] table [
                col sep=comma,
                mark=none
            ]{Data/Figure05_1/Figure05_1_x.csv};

            \addplot[tumOrange, line width=1.2pt] table [
                col sep=comma,
                mark=none
            ]{Data/Figure05_1/Figure05_1_z.csv};

            \addplot[tumYellow, line width=1.2pt] table [
                col sep=comma,
                mark=none
            ]{Data/Figure05_1/Figure05_1_phi.csv};

            \addplot[tumPink, line width=1.2pt] table [
                col sep=comma,
                mark=none
            ]{Data/Figure05_1/Figure05_1_gamma.csv};

            \addplot[tumGreen, line width=1.2pt] table [
                col sep=comma,
                mark=none
            ]{Data/Figure05_1/Figure05_1_l.csv};

            \path[name path=axis] (0,-10) -- (0,10);
            \path[name path=f] (0.9,-10) -- (0.9,10);
            \addplot[
                color=gray,
                fill opacity=0.2
            ] fill between[
                of=axis and f
            ];
            
        \end{axis}

        \begin{axis}[
            name=second plot,
            at={(main plot.below south west)},
            yshift=-0.1cm,
            anchor=north west,
            xtick={0,.5,1,1.5},
            ytick={-.6,-.4,-.2,0,.2,.4,.6,.8,1},
            xmin=0,
            xmax=1.9,
            ymin=-.7,
            ymax=1.1,
            legend pos=outer north east,
            grid = major,
            legend entries={$\Dot{x}$,$\Dot{z}$,$\Dot{\phi}$,$\Dot{\alpha}$,$\Dot{l}$},
            width=7cm,
            height=3.5cm
        ]

            \addplot[tumBlue, line width=1.2pt] table [
                col sep=comma,
                mark=none
            ]{Data/Figure05_2/Figure05_2_dx.csv};

            \addplot[tumOrange, line width=1.2pt] table [
                col sep=comma,
                mark=none
            ]{Data/Figure05_2/Figure05_2_dz.csv};

            \addplot[tumYellow, line width=1.2pt] table [
                col sep=comma,
                mark=none
            ]{Data/Figure05_2/Figure05_2_dphi.csv};

            \addplot[tumPink, line width=1.2pt] table [
                col sep=comma,
                mark=none
            ]{Data/Figure05_2/Figure05_2_dgamma.csv};

            \addplot[tumGreen, line width=1.2pt] table [
                col sep=comma,
                mark=none
            ]{Data/Figure05_2/Figure05_2_dl.csv};

            \path[name path=axis] (0,-10) -- (0,10);
            \path[name path=f] (0.9,-10) -- (0.9,10);
            \addplot[
                color=gray,
                fill opacity=0.2
            ] fill between[
                of=axis and f
            ];
            
        \end{axis}

        \begin{axis}[
            at={(second plot.below south west)},
            yshift=-0.1cm,
            anchor=north west,
            xlabel={Time $[\sqrt{l_0 / g}]$},
            xtick={0,.5,1,1.5},
            ytick={-.4,-.2,0,.2,.4,.6},
            xmin=0,
            xmax=1.9,
            ymin=-.5,
            ymax=.6,
            legend pos=outer north east,
            grid = major,
            legend entries={$u_{\alpha}$,$u_\mathrm{l}$},
            width=7cm,
            height=3.5cm
        ]

            \addplot+[jump mark left, tumPink, line width=1.2pt, mark size=1pt] table [
                col sep=comma
            ]{Data/Figure05_3/Figure05_3_ugamma.csv};

            \addplot+[jump mark left, tumGreen, line width=1.2pt, mark size=1pt, mark=*] table [
                col sep=comma
            ]{Data/Figure05_3/Figure05_3_ul.csv};

            \path[name path=axis] (0,-10) -- (0,10);
            \path[name path=f] (0.9,-10) -- (0.9,10);
            \addplot[
                color=gray,
                fill opacity=0.2
            ] fill between[
                of=axis and f
            ];
            
        \end{axis}
    \end{tikzpicture}
    \end{minipage}
    \caption{Actuated hopping forward gait of the prismatic monopod as result of homotopic continuation. The cost function is $c = \boldsymbol{\xi}\T \boldsymbol{\xi}$. The gray area marks the stance phase. The continuous states and velocities are similar to the respective quasi-passive gait but show small differences, such as in the absolute angle of the main body. The plot on the bottom shows the actuation profile of the parallel elastic actuators for each grid point.}
    \label{fig:ActPrismUSq}
\end{figure}

\subsection{Actuated Gait}
We take the quasi-passive gait $(\varepsilon = 0)$ obtained from root-search with direct collocation and apply continuation on the first-order optimality until we find an actuated gait $(\varepsilon = 1)$ including the actuation profile $\bxi \neq \mathbf{0}$. Note that the virtual energy injection parameter $\gamma$ is constant throughout the procedure. The virtual energy injection is canceled out with the continuation variable $\varepsilon \rightarrow 1$. The operating point $\bar{O}$ continues to define the respective gait over the continuation. The control input for the actuation of the prismatic monopod is the force $u_\mathrm{l}$ acting on the leg length $l$ and the torque $u_\alpha$ acting on the joint angle $\alpha$. For homotopic continuation, we define the  actuator forces squared as cost function: $c = \boldsymbol{\xi}\T \boldsymbol{\xi}$.\par
\begin{figure}
    \centering
    \pgfplotsset{footnotesize}
    \begin{tikzpicture}
        \begin{axis}[
            title={\rev{Prismatic Monopod: Homotopic Continuation}},
            xlabel={$\varepsilon$},
            xtick={0,.2,.4,.6,.8,1},
            ytick={0,.5,1,1.5},
            xmin=0,
            xmax=1,
            ymin=0,
            ymax=1.7,
            legend pos=outer north east,
            grid = major,
            legend entries={$t_{\text{stance}}$,$t_{\text{flight}}$,$c$,$\mu_{\text{min}}$},
            width=7cm,
            height=3cm
        ]

        \addplot[tumBlue, line width=1.2pt, no markers] table [
            col sep=comma
        ]{Data/Figure05_4/Figure05_4_tS.csv};

        \addplot[tumBlue, line width=1.2pt, no markers, dashed] table [
            col sep=comma
        ]{Data/Figure05_4/Figure05_4_tF.csv};

        \addplot[tumLighterBlue, line width=1.2pt, no markers, dashed] table [
            col sep=comma
        ]{Data/Figure05_4/Figure05_4_cost.csv};

        \addplot[tumLighterBlue, line width=1.2pt, no markers] table [
            col sep=comma
        ]{Data/Figure05_4/Figure05_4_eigH.csv};
            
        \end{axis}
    \end{tikzpicture}
    \caption{Evolving values over the homotopic continuation of the hopping forward gait of the prismatic monopod with cost $c = \boldsymbol{\xi}\T \boldsymbol{\xi}$. The values correspond to a quasi-passive solution at $\varepsilon = 0$ and evolve to the actuated solution at $\varepsilon = 1$. The flight phase duration $t_{\text{flight}}$ decreases over the continuation. The cost \revrm{function }$c$ \rev{starts from zero at the quasi-passive solution and increases over the continuation} to compensate for the missing virtual energy injection. \rev{This is contrary to other optimization approaches, where the cost decreases from non-optimal values to its optimum, while in our approach gaits are always optimal.} The smallest eigenvalue $\mu_{\text{min}}$ of the second-order optimality condition is always positive, which proves that the solutions are always a minimum.} 
    \label{fig:ContinPrismUSq}
\end{figure}
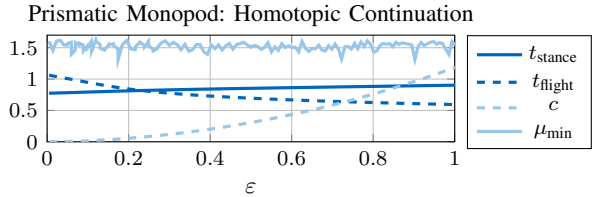

Figure \ref{fig:ContinPrismUSq} shows how the problem behaves over the continuation with $\varepsilon = 0 \rightarrow 1$. The cost function $c$ starts at zero, as there is zero actuation at the quasi-passive solution. With increasing $\varepsilon$ and the adapting actuation profile, the cost function is forced to increase to values $c > 0$. When reaching the state of no virtual energy injection at $\varepsilon = 1$, the actuation profile has changed its values to support the motion of the prismatic monopod optimally with respect to the cost function~$c$, such that it can perform the desired forward gait at average speed $\bar{\dot{x}} = 0.3 \sqrt{l_0g}$. \revrm{Note, how the flight phase duration $t_{\text{flight}}$ decreases over the continuation. }The actuated solution, which primarily inserts energy into the system during the stance phase (by pushing off the ground), consequently results in an extended stance phase. Figure \ref{fig:ActPrismUSq} depicts the derived actuation profile and the approximated active gait on the prismatic monopod. \revrm{Note, that over the continuation also the initial values $\x_0$ of a stride change. }In Figure \ref{fig:ContinPrismUSq} we can observe that the second-order optimality condition \eqref{eq:SecondOrdOpt} is fulfilled over the whole continuation, since its smallest eigenvalue is positive $\mu_{\text{min}} > 0 \,\forall \varepsilon$, i.e., the solution is minimal.

\section{Example: Sagittal Quadruped}
\label{sec:Quadru}
To cover systems with increasing complexity we apply a sagittal quadruped model with $n_\mathrm{q} = 7$. The general coordinates of the system are $\mathbf{q} = [x,z,\phi,\alpha_{\mathrm{hu}},\alpha_{\mathrm{hl}},\alpha_{\mathrm{fu}},\alpha_{\mathrm{fl}}]\T$. The sagittal quadruped model uses series elastic actuation; thus the control inputs are the motor positions $\mathbf{u} = [u_\mathrm{hu},u_\mathrm{hl},u_\mathrm{fu},u_\mathrm{fl}]$. The model structure is depicted in Figure~\ref{fig:SagittalQuadruped}.

\begin{figure}
    \label{fig:SagittalQuadruped}
    \centering
    \resizebox{0.7\linewidth}{!}{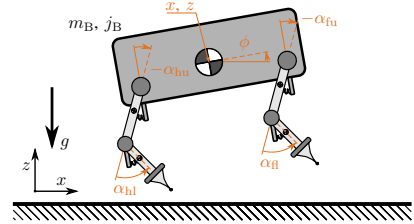}
    \caption{The sagittal quadruped model contains a main body with mass $m_\mathrm{B}$ and inertia $j_\mathrm{B}$, and two legs (hind and front) consisting of an upper and lower leg, as well as a foot as point mass. In each leg there are two radial springs with stiffness and damping at each the hip and the knee joint. The configuration of the system is described by $\mathbf{q} = [x\,z\,\phi\,\alpha_{\mathrm{hu}}\,\alpha_{\mathrm{hl}}\,\alpha_{\mathrm{fu}}\,\alpha_{\mathrm{fl}}]\T$.}
\end{figure}

\subsection{Quasi-Passive Gait}
\rev{The desired gait for this system is bounding forward, which results in a sequence of four different phases. The number of grid points per phase is set to $N = 10$. We start the root-search with a fixed energy level $E(\x(t=0))=1.3\,m_0gl_0$ as operating point, and a freely chosen motion as initial guess with phase sequence \textit{touch-down front}$\rightarrow$\textit{lift-off hind}$\rightarrow$\textit{lift-off front}$\rightarrow$\textit{touch-down hind}. Figure~\ref{fig:SnapshotPrism} includes snapshots of the resulting motion.}

\subsection{Actuated Gait}
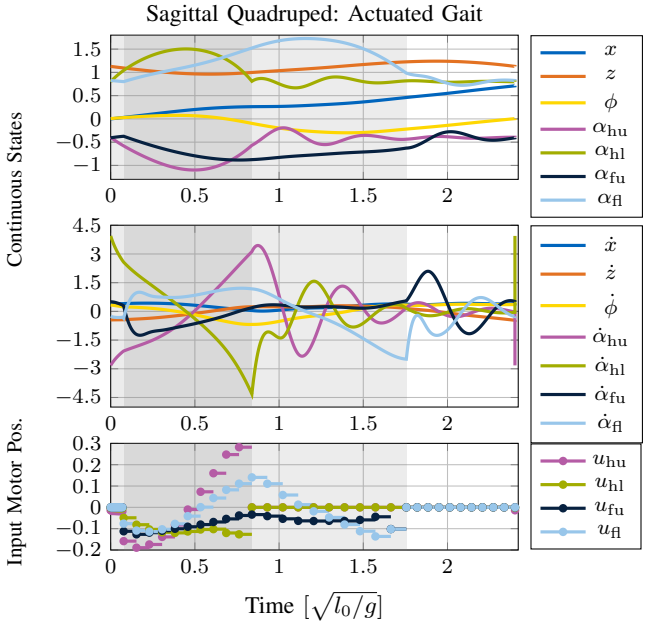
\begin{figure}
\begin{minipage}{0.05\linewidth}
    \rotatebox{90}{\footnotesize Input Motor Pos. \hspace{1.8cm} Continuous States \hspace{0.7cm}}
\end{minipage}
\begin{minipage}{0.95\linewidth}
    \pgfplotsset{footnotesize}
    \begin{tikzpicture}
        \begin{axis}[
            name=main plot,
            title={\rev{Sagittal Quadruped: Actuated Gait}},
            xtick={0,.5,1,1.5,2},
            ytick={-1,-.5,0,.5,1,1.5},
            xmin=0,
            xmax=2.42,
            ymin=-1.3,
            ymax=1.8,
            legend pos=outer north east,
            grid = major,
            legend entries={$x$,$z$,$\phi$,$\alpha_{\mathrm{hu}}$,$\alpha_{\mathrm{hl}}$,$\alpha_{\mathrm{fu}}$,$\alpha_{\mathrm{fl}}$},
            width=7cm,
            height=3.5cm
        ]

            \addplot[tumBlue, line width=1.2pt] table [
                col sep=comma,
                mark=none
            ]{Data/Biped_1/Biped_1_x.csv};

            \addplot[tumOrange, line width=1.2pt] table [
                col sep=comma,
                mark=none
            ]{Data/Biped_1/Biped_1_z.csv};

            \addplot[tumYellow, line width=1.2pt] table [
                col sep=comma,
                mark=none
            ]{Data/Biped_1/Biped_1_phi.csv};

            \addplot[tumPink, line width=1.2pt] table [
                col sep=comma,
                mark=none
            ]{Data/Biped_1/Biped_1_gammaUpL.csv};

            \addplot[tumGreen, line width=1.2pt] table [
                col sep=comma,
                mark=none
            ]{Data/Biped_1/Biped_1_gammaLoL.csv};

            \addplot[tumDarkerBlue, line width=1.2pt] table [
                col sep=comma,
                mark=none
            ]{Data/Biped_1/Biped_1_gammaUpR.csv};

            \addplot[tumLighterBlue, line width=1.2pt] table [
                col sep=comma,
                mark=none
            ]{Data/Biped_1/Biped_1_gammaLoR.csv};

            \path[name path=axis] (0.0,-10) -- (0.0,10);
            \path[name path=f] (0.08,-10) -- (0.08,10);
            \addplot[
                color=gray,
                fill opacity=0.15
            ] fill between[
                of=axis and f
            ];

            \path[name path=axis] (0.08,-10) -- (0.08,10);
            \path[name path=f] (0.84,-10) -- (0.84,10);
            \addplot[
                color=gray,
                fill opacity=0.3
            ] fill between[
                of=axis and f
            ];

            \path[name path=axis] (0.84,-10) -- (0.84,10);
            \path[name path=f] (1.76,-10) -- (1.76,10);
            \addplot[
                color=gray,
                fill opacity=0.15
            ] fill between[
                of=axis and f
            ];
            
        \end{axis}

        \begin{axis}[
            name=second plot,
            at={(main plot.below south west)},
            yshift=-0.1cm,
            anchor=north west,
            xtick={0,.5,1,1.5,2},
            ytick={-4.5,-3,-1.5,0,1.5,3,4.5},
            xmin=0,
            xmax=2.42,
            ymin=-5,
            ymax=4.5,
            legend pos=outer north east,
            grid = major,
            legend entries={$\Dot{x}$,$\Dot{z}$,$\Dot{\phi}$,$\Dot{\alpha}_{\mathrm{hu}}$,$\Dot{\alpha}_{\mathrm{hl}}$,$\Dot{\alpha}_{\mathrm{fu}}$,$\Dot{\alpha}_{\mathrm{fl}}$},
            width=7cm,
            height=4cm
        ]

            \addplot[tumBlue, line width=1.2pt] table [
                col sep=comma,
                mark=none
            ]{Data/Biped_2/Biped_2_dx.csv};

            \addplot[tumOrange, line width=1.2pt] table [
                col sep=comma,
                mark=none
            ]{Data/Biped_2/Biped_2_dz.csv};

            \addplot[tumYellow, line width=1.2pt] table [
                col sep=comma,
                mark=none
            ]{Data/Biped_2/Biped_2_dphi.csv};

            \addplot[tumPink, line width=1.2pt] table [
                col sep=comma,
                mark=none
            ]{Data/Biped_2/Biped_2_dgammaUpL.csv};

            \addplot[tumGreen, line width=1.2pt] table [
                col sep=comma,
                mark=none
            ]{Data/Biped_2/Biped_2_dgammaLoL.csv};

            \addplot[tumDarkerBlue, line width=1.2pt] table [
                col sep=comma,
                mark=none
            ]{Data/Biped_2/Biped_2_dgammaUpR.csv};

            \addplot[tumLighterBlue, line width=1.2pt] table [
                col sep=comma,
                mark=none
            ]{Data/Biped_2/Biped_2_dgammaLoR.csv};

            \path[name path=axis] (0.0,-10) -- (0.0,10);
            \path[name path=f] (0.08,-10) -- (0.08,10);
            \addplot[
                color=gray,
                fill opacity=0.15
            ] fill between[
                of=axis and f
            ];

            \path[name path=axis] (0.08,-10) -- (0.08,10);
            \path[name path=f] (0.84,-10) -- (0.84,10);
            \addplot[
                color=gray,
                fill opacity=0.3
            ] fill between[
                of=axis and f
            ];

            \path[name path=axis] (0.84,-10) -- (0.84,10);
            \path[name path=f] (1.76,-10) -- (1.76,10);
            \addplot[
                color=gray,
                fill opacity=0.15
            ] fill between[
                of=axis and f
            ];
            
        \end{axis}

        \begin{axis}[
            at={(second plot.below south west)},
            yshift=-0.1cm,
            anchor=north west,
            xlabel={Time $[\sqrt{l_0 / g}]$},
            xtick={0,.5,1,1.5,2},
            ytick={-.2,-.1,0,.1,.2,.3},
            xmin=0,
            xmax=2.42,
            ymin=-.2,
            ymax=.3,
            legend pos=outer north east,
            grid = major,
            legend entries={$u_{\mathrm{hu}}$,$u_{\mathrm{hl}}$,$u_{\mathrm{fu}}$,$u_{\mathrm{fl}}$},
            width=7cm,
            height=3cm
        ]

            \addplot+[jump mark left, tumPink, line width=1.2pt, mark size=1pt] table [
                col sep=comma
            ]{Data/Biped_3/Biped_3_uUpL.csv};

            \addplot+[jump mark left, tumGreen, line width=1.2pt, mark size=1pt, mark=*] table [
                col sep=comma
            ]{Data/Biped_3/Biped_3_uLoL.csv};

            \addplot+[jump mark left, tumDarkerBlue, line width=1.2pt, mark size=1pt] table [
                col sep=comma
            ]{Data/Biped_3/Biped_3_uUpR.csv};

            \addplot+[jump mark left, tumLighterBlue, line width=1.2pt, mark size=1pt, mark=*] table [
                col sep=comma
            ]{Data/Biped_3/Biped_3_uLoR.csv};

            \path[name path=axis] (0.0,-10) -- (0.0,10);
            \path[name path=f] (0.08,-10) -- (0.08,10);
            \addplot[
                color=gray,
                fill opacity=0.15
            ] fill between[
                of=axis and f
            ];

            \path[name path=axis] (0.08,-10) -- (0.08,10);
            \path[name path=f] (0.84,-10) -- (0.84,10);
            \addplot[
                color=gray,
                fill opacity=0.3
            ] fill between[
                of=axis and f
            ];

            \path[name path=axis] (0.84,-10) -- (0.84,10);
            \path[name path=f] (1.76,-10) -- (1.76,10);
            \addplot[
                color=gray,
                fill opacity=0.15
            ] fill between[
                of=axis and f
            ];
            
        \end{axis}
    \end{tikzpicture}
    \end{minipage}
    \rev{\caption{Actuated bounding gait of the sagittal quadruped as result of homotopic continuation. The cost function is $c = \boldsymbol{\xi}\T \boldsymbol{\xi}$. The light gray areas mark single support, while the dark gray area marks double support. On the bottom, the motor trajectory is depicted. The actuation allows only non-zero positions, if the respective leg is in contact with the ground.} \label{fig:ActQuadru}}
\end{figure}

\rev{An extension onto multi-legged systems does not alter the homotopic continuation framework. We continue to use one single energy injection parameter $\gamma$ and one single homotopy parameter $\varepsilon$. We choose the cost function $c = \bxi\T\bxi$ and apply continuation until $\varepsilon = 1$. Note, that we have set additional constraints, such that the motors can only set non-zero positions, if the respective leg is in contact with the ground. This avoids unnecessary motion of the legs in flight. The obtained actuated gait without virtual energy injection is shown in Figure~\ref{fig:ActQuadru}.}

\section{Conclusion}
\label{sec:Conclusions}
We present a new, modular approach to obtain quasi-passive gaits for dissipative systems that are not dependent on complex actuation profiles, but on one single energy input parameter. The underlying structure consists of a root-finding framework in combination with virtual energy injection. This energy injection only depends on the velocities of the system and is thus similar to a positive damping force, which acts like an external force on the robot. A passive gait search algorithm is complemented by an effective way to derive actuation profiles for the real dissipative system that is no longer dependent on virtual energy injection. This approach uses continuation and an input homotopy to convert the virtually supported passive gait into an actuated gait with an actuation profile of arbitrary shape. A cost function is defined for the actuation derivation, which influences the resulting gait and motor profile. The results validate the concept \rev{on different systems with increasing complexity.}\par
The proposed approach is able to bypass a variety of problems associated with simplified modeling in legged locomotion. Energy losses caused by impacting foot masses or high-frequency oscillations in non-dissipative systems, as it might occur in \cite{raff2022generating}, pose no problem in our framework. Instead, we keep dissipative terms and counteract with virtual energy injection scaled with a single parameter, regardless of the system's complexity \rev{or its number of legs}. This constitutes a central advantage over optimal control, for which large NLPs need to be defined and solved \cite{Wensing2024}. As quasi-passive gaits are identified rapidly, the shape of the gait can be examined and modified in an early stage of the approach. Virtual energy injection of the form \eqref{eq:EnrgyInjPMC} enables identification of efficient quasi-passive motions close to the natural frequencies of the system \cite{krack2015nonlinear}. The continuation into a deployable active gait without the global search contributes to the approach's simplicity. \rev{This is possible for both single- and multi-legged systems using one homotopy parameter.} The \rev{resulting} actuated solution remains close to the initial quasi-passive gait.\par
There are multiple items for which further investigation in future work will be worthwhile. For the virtual energy injection, \rev{a specific} kind of injection was chosen out of many possibilities. This choice certainly influences the shape\rev{, stability, robustness and other properties} of resulting quasi-passive gaits and their associated active gait. \rev{For future works we want to explore different injection possibilities and put one to another into relation via continuation. }\revrm{ While different kinds were tested, it is still possible to investigate the respective advantages more deeply. Furthermore, continued research on homotopic continuation in bipeds is essential for obtaining motor trajectories for bipedal running.} \rev{Furthermore, for} proper bipedal motion, inequality constraints must be included in the homotopic continuation approach. Another significant challenge is the implementation of a stabilizing controller, which is essential for applying the identified optimal active gaits to hardware.
\bibliographystyle{IEEEtran}
%



\bibliography{bibtex/bib/paper}

%









\title{Supplement to RA-L Publication: Discovering Optimal Natural Gaits \\ of Dissipative Systems via Virtual Energy Injection}

\author{Korbinian~Griesbauer$^{1}$,
        Davide~Calzolari$^{1}$,
        Maximilian~Raff$^{2}$,
        C.~David~Remy$^{2}$,
        Alin~Albu-Sch\"{a}ffer$^{1}$
\thanks{$^{1}$K. Griesbauer, D. Calzolari, and A. Albu-Sch\"{a}ffer are with School of Computation, Information and Technology, Technical University of Munich (TUM), 85748 Garching, Germany, and with the Institute of Robotics and Mechatronics, German Aerospace Center (DLR), 82234 Weßling, Germany {\fontfamily{qcr}\selectfont \{korbinian.griesbauer,davide.calzolari,alin.albu-schaeffer\}@dlr.de}}
\thanks{$^{2}$M. Raff and C.~D. Remy are with the Institute for Adaptive Mechanical Systems, University of Stuttgart, 70569 Stuttgart, Germany {\tt\footnotesize \{maximilian.raff,david.remy\}@iams.uni-stuttgart.de}}%
}

\markboth{arXiv, November~2025}%
{Griesbauer \MakeLowercase{\textit{et al.}}: Supplement: Discovering Optimal Natural Gaits of Dissipative Systems}


\maketitle

\begin{abstract}
This article acts as supplementary material for the IEEE Robotics and Automation Letters publication with the title ``Discovering Optimal Natural Gaits of Dissipative Systems via Virtual Energy Injection''.
This paper presents the detailed mathematical formulation of our approach, along with implementation details and simulation results, which were omitted from the Letters publication due to space constraints.
\end{abstract}

\section{Introduction}
We structure this article as follows: In Section \ref{sec:EoMs} we provide a detailed description of the shown models. This includes equations of motions (EoM), constraints, event functions and mapping between phases. Each subsection is dedicated to one model. We also include the segmented monopod model, which is not entirely described in the main document. In Section \ref{sec:root} we show how exactly we apply direct collocation and implement it in MATLAB and CasADi \cite{Andersson2019}. Section \ref{sec:contin} gives an overview of the continuation approach and shows results of the segmented monopod model.

\section{Equations of Motion}
\label{sec:EoMs}
In the following, we present the EoM of the examples.

\subsection{Prismatic Monopod}
The model is composed of a main body with mass~$m_\mathrm{B}$ and inertia~$j_\mathrm{B}$. The variables~$x$ and~$z$ describe the horizontal and vertical positions of the CoM, respectively. The main body's pitch angle is denoted by~$\phi$. The leg is mounted on a rotational joint positioned on the main body's CoM. The joint angle between the main body and leg is~$\alpha$. The two parts are connected with a torsional spring with stiffness~$k_\mathrm{H}$ and damping~$d_\mathrm{H}$. The spring is uncompressed for $\alpha = 0$. The leg has mass~$m_\mathrm{L}$ and inertia~$j_\mathrm{L}$ and is connected to a foot with a translational linear spring. The foot is at radial position~$l$ relative to the CoM of the main body. The generalized coordinates are thus $\mathbf{q} = [x, z, \phi,\alpha,l]\T$, with $n_\mathrm{q} = 5$. The translational spring is uncompressed at $l = l_0$ and has stiffness~$k_\mathrm{L}$ and damping~$d_\mathrm{L}$. Parallel to both springs, there are actuators that produce the torques $\mathbf{u} = [u_\alpha,u_\mathrm{l}]\T$. The foot is modeled as a point mass with mass~$m_\mathrm{F}$. Figure \ref{fig:PrismaticMonopod} shows the structure of the prismatic monopod model. Table~\ref{tab:parPrism} contains the values of all parameters.\par

\begin{figure}
    \centering
    \resizebox{0.7\linewidth}{!}{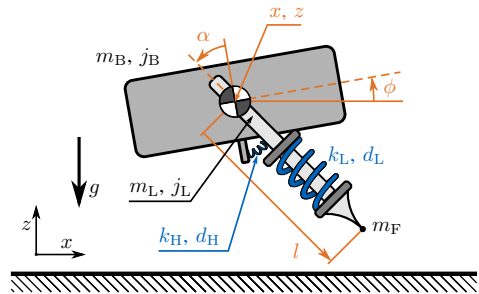}
    \caption{The prismatic monopod model contains a main body with mass $m_\mathrm{B}$ and inertia $j_\mathrm{B}$, a leg with mass $m_\mathrm{L}$ and inertia $j_\mathrm{L}$ and a foot as point mass $m_\mathrm{F}$. There is a radial spring with stiffness $k_\mathrm{H}$ and damping $d_\mathrm{H}$ at the hip joint, as well as a linear spring with stiffness $k_\mathrm{L}$ and damping $d_\mathrm{L}$ between the leg and foot. The configuration of the system is described by $\mathbf{q} = [x\,z\,\phi\,\alpha\,l]\T$.}
    \label{fig:PrismaticMonopod}
\end{figure}

We derive the EoM using Lagrangian mechanics. First, we define the potential energy of each part (base~$\mathrm{B}$, leg~$\mathrm{L}$, foot~$\mathrm{F}$) and of the springs~$\mathrm{S}$:
\begin{subequations}
        \begin{align}
                V_\mathrm{B} &= m_\mathrm{B} g z \\
                V_\mathrm{L} &= m_\mathrm{L} g z \\
                V_\mathrm{F} &= m_\mathrm{F} g \big(z - l \cos(\phi + \alpha)\big) \\
                V_\mathrm{S} &= \frac{1}{2} \big( k_\mathrm{H} (-\alpha)^2 + k_\mathrm{L} (l_0 - l)^2 \big)
        \end{align}
\end{subequations}
The kinetic energy of each part is given by
\begin{subequations}
        \begin{align}
                T_\mathrm{B} &= \frac{1}{2} \big( m_\mathrm{B} \dot{x}^2 + m_\mathrm{B} \dot{z}^2 + j_\mathrm{B} \dot{\phi}^2 \big) \\
                T_\mathrm{L} &= \frac{1}{2} \big( m_\mathrm{L} \dot{x}^2 + m_\mathrm{L} \dot{z}^2 + j_\mathrm{L} (\varkappa)^2 \big) \\
                \begin{split}
                        T_\mathrm{F} = &\frac{1}{2} m_\mathrm{F} \big( \dot{x} + \sin(\varkappa) \dot{l} + \cos(\varkappa) \varkappa l \big)^2 + \\
                        &\frac{1}{2} m_\mathrm{F} \big( \dot{z} - \cos(\varkappa) \dot{l} + \sin(\varkappa) \varkappa l \big)^2
                \end{split}
        \end{align}
\end{subequations}
where $\varkappa = \dot{\phi} + \dot{\alpha}$. We define the Lagrangian
\begin{equation}
        \mathrm{Lag} = \sum T - \sum V
        \label{eq:Lagrangian}
\end{equation}
and the EoM of the system
\begin{equation}
        \frac{d}{dt} \bigg( \frac{\partial \mathrm{Lag}}{\partial \dq} \bigg) + \frac{\partial \mathrm{Lag}}{\partial \q} = \mathbf{0} .
        \label{eq:LagrEoM}
\end{equation}
We can bring the given EoM into the form
\begin{equation}
        \mathbf{M}(\q) \ddot{\mathbf{q}} = \mathbf{n}(\q,\dq).
        \label{eq:EoMwoCont}
\end{equation}
In the case of the prismatic monopod we have parallel actuation. Thus, the actuation is added to the EoM as given torques $u_\alpha$ and $u_l$. Note that the virtual input term $\mathbf{f}_\mathrm{E}$ is added to the EoM as in the main document, Eq. (12). For the computation of the constraint Jacobian $\mathbf{W}_i$ we need to define constraints for each phase. 
\begin{subequations}
        \begin{align}
                \mathbf{g}_1 &= 0 \\
                \mathbf{g}_2 &= \begin{bmatrix}
                        x +l \sin(\phi+\alpha) \\
                        z -l \cos(\phi+\alpha)
                \end{bmatrix}
        \end{align}
\end{subequations}
The constraint Jacobian is given by $\mathbf{W}_i \T := \partial \mathbf{g}_i / \partial \q$. Table~\ref{tab:phasePrism} shows the numbering of the phases, as well as the event functions. Next, we define the event functions necessary for the hybrid dynamics 
\begin{subequations}
        \begin{align}
                e_1^2(\x) &= - z + l \sin(\phi + \alpha)\\
                e_2^1(\x) &= - \uplambda_{2,2}
        \end{align}
\end{subequations}
where $\uplambda_{2,2}$ is the entry of the contact forces in $z$-direction:
\begin{equation}
        \boldsymbol{\uplambda}_2 = \big( \mathbf{W}_2\T (\mathbf{M}^{-1} \mathbf{W}_2) \big)^{-1} \big( -\mathbf{W}_2\T (\mathbf{M}^{-1}\mathbf{n}) - \dot{\mathbf{W}}\T\dq \big)
        \label{eq:contact}
\end{equation}
If an event function crosses zero in positive direction, the event is triggered. We project from phase $i$ to $j$ with a discrete map
\begin{equation}\label{eq:discrete}
    \boldsymbol{\Delta}_i^j(\mathbf{x}) = \begin{bmatrix}
        \mathbf{q} \\
        \big(\mathbf{I} - \mathbf{M}^{-1} \mathbf{W}_j(\mathbf{W}_j\T \mathbf{M}^{-1} \mathbf{W}_j )^{-1} \mathbf{W}_j\T \big) \, \dot{\mathbf{q}}
    \end{bmatrix}.
\end{equation}
Thus, we can map the states right before and after the event~$e_i^j$ with $\mathbf{x}^+ = \boldsymbol{\Delta}_i^j(\mathbf{x}^-)$ \cite{brogliato2016nonsmooth}.

\begin{table}
        \centering
        \begin{tabular}{c l}
                \thickhline
                \textbf{no.} & \textbf{phase} \\
                \hline
                1 & flight \\
                2 & stance \\
                \thickhline
                \textbf{event function} & \textbf{event} \\
                \hline
                $e_1^2$ & touch-down \\
                $e_2^1$ & lift-off \\
                \thickhline 
        \end{tabular}
        \caption{Phases and events of the monopod models.}
        \label{tab:phasePrism}
\end{table}

\begin{table}
\centering
    \begin{tabular}{l l c c}
        
        \thickhline
        Parameter &  & Value & Unit\\
        \hline
        total mass & $m_0$ & 1 & $\cdot$ \\
        uncompressed leg length & $l_0$ & 1 & $\cdot$ \\
        gravitational constant & $g$ & 1 & $\cdot$ \\
        mass main body & $m_\mathrm{B}$ & 0.7 & $m_0$ \\
        mass leg & $m_\mathrm{L}$ & 0.2 & $m_0$ \\
        mass foot & $m_\mathrm{F}$ & 0.1 & $m_0$ \\
        inertia main body & $j_\mathrm{B}$ & 0.4 & $m_0 l_0^2$ \\
        inertia leg & $j_\mathrm{L}$ & 0.004 & $m_0 l_0^2$ \\
        stiffness rotational spring & $k_\mathrm{H}$ & 1 & $m_0 g l_0/\text{rad}$ \\
        damping rotational spring & $d_\mathrm{H}$ & 0.02 & $m_0 \sqrt{l_0^3 g^3}/\text{rad}$ \\
        stiffness linear spring & $k_\mathrm{L}$ & 20 & $m_0 g/l_0$ \\
        damping linear spring & $d_\mathrm{L}$ & 0.85 & $m_0 \sqrt{g/l_0}$ \\
        \thickhline
        
    \end{tabular}
    \caption{Parameters of the Prismatic Monopod Model.}
    \label{tab:parPrism}
\end{table}

\subsection{Segmented Monopod}
\begin{figure}
    \centering
    \resizebox{0.7\linewidth}{!}{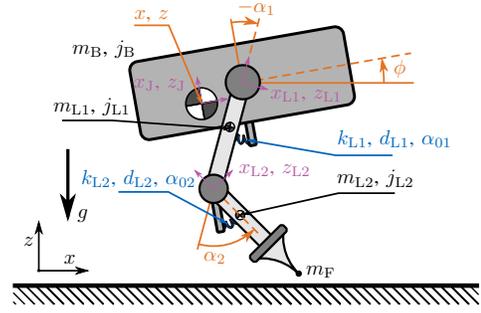}
    \caption{The segmented monopod model consists of a main body with mass $m_\mathrm{B}$ and inertia $j_\mathrm{B}$, an upper leg with mass $m_\mathrm{L1}$ and inertia $j_\mathrm{L1}$, a lower leg with mass $m_\mathrm{L2}$ and inertia $j_\mathrm{L2}$, and a foot as point mass~$m_\mathrm{F}$. The positions of the CoMs of each segment are defined relative to the segment's joint. There are two rotational springs with stiffness $k_\mathrm{L1}, k_\mathrm{L2}$ and damping $d_\mathrm{L1}, d_\mathrm{L2}$ at joint 1 and 2, respectively. Both springs 1 and 2 are uncompressed at $\alpha_{01}, \alpha_{02}$ relative to the vertical of the main body. The configuration of the system is described by $\mathbf{q} = [x\,z\,\phi\,\alpha_1\,\alpha_2]\T$.}
    \label{fig:SegmentedMonopod}
\end{figure}

In nature, most legged animals developed articulated legs for increased versatility. We apply a segmented monopod model, which is comparable to legs of this type. It consists of a main body with mass $m_\mathrm{B}$ and inertia $j_\mathrm{B}$.
The horizontal and vertical positions of its CoM and the pitch angle of the body are again $x$, $z$, and $\phi$. The upper leg is attached to the main body with joint angle $\alpha_1$.
The torsional spring connecting both parts has stiffness $k_1$ and damping $d_1$.
The angle at which the spring is uncompressed is defined as $\alpha_{01}$ relative to a vertical through the main body. The upper leg has mass $m_\mathrm{L1}$, inertia $j_\mathrm{L1}$, and length $l_1$. The lower leg is connected to the main body with a torsional linear spring with stiffness $k_2$ and damping $d_2$.
At angle $\alpha_{02}$ relative to a vertical through the main body the spring is uncompressed, just as the first torsional spring.
The foot is a point mass $m_\mathrm{F}$.
The general coordinates of the system are $\mathbf{q} = [x ,z ,\phi ,\alpha_1 ,\alpha_2]\T$, with $n_\mathrm{q} = 5$.
There are actuators mounted in series with the springs with motor positions $\mathbf{u} = [u_{\alpha_1} ,u_{\alpha_2}]\T$. Figure~\ref{fig:SegmentedMonopod} shows the structure of the segmented monopod model. Again, we normalize all values with respect to $m_0$, $g$, and $l_0 = l_1+l_2$. Table \ref{tab:parSegm} contains the values of all parameters of the model.\par

We can note the position vectors of each part of the system (base~$\mathrm{B}$, upper leg~$\mathrm{L}1$, lower leg~$\mathrm{L}2$, foot~$\mathrm{F}$, springs~$\mathrm{S}$):
\begin{subequations}
        \begin{align}
                \mathbf{r}_\mathrm{B} &= [x, z]\T \\
                \mathbf{r}_{\mathrm{L}1} &= \begin{bmatrix}
                        x + x_\mathrm{J} c_\phi - z_\mathrm{J} s_\phi + x_{\mathrm{L}1} c_{\phi\alpha_1} - z_{\mathrm{L}1} s_{\phi\alpha_1} \\
                        z + x_\mathrm{J}s_\phi + z_\mathrm{J}c_\phi + x_{\mathrm{L}1} s_{\phi\alpha_1} + z_{\mathrm{L}1} c_{\phi\alpha_1}
                \end{bmatrix} \\
                \mathbf{r}_{\mathrm{L}2} &= \begin{bmatrix}
                        x + x_\mathrm{J}c_\phi - z_\mathrm{J}s_\phi + l_1s_{\phi\alpha_1} + \\ +x_{\mathrm{L}2} c_{\phi\alpha_1\alpha_2} - z_{\mathrm{L}2} s_{\phi\alpha_1\alpha_2} \\ z + x_\mathrm{J}s_\phi + z_\mathrm{J}c_\phi - l_1c_{\phi\alpha_1} + \\ + x_{\mathrm{L}2} s_{\phi\alpha_1\alpha_2} + z_{\mathrm{L}2} c_{\phi\alpha_1\alpha_2}
                \end{bmatrix} \\
                \mathbf{r}_\mathrm{F} &= \begin{bmatrix}
                        x + x_\mathrm{J} c_\phi - z_\mathrm{J} s_\phi + l_1 s_{\phi\alpha_1} + l_2 s_{\phi\alpha_1\alpha_2} \\
                        z + x_\mathrm{J} s_\phi + z_\mathrm{J} c_\phi - l_1 c_{\phi\alpha_1} - l_2 c_{\phi\alpha_1\alpha_2}
                \end{bmatrix}
        \end{align}
\end{subequations}
where $c_{ij}$ and $s_{ij}$ denote $\cos(i+j)$ and $\sin(i+j)$, respectively. Again, we define the potential energy of the system with help of the position vectors:
\begin{subequations}
        \begin{align}
                V_\mathrm{B} &= m_\mathrm{B}gr_{\mathrm{B},2} \\
                V_{\mathrm{L}1} &= m_{\mathrm{L}1}gr_{\mathrm{L}1,2} \\
                V_{\mathrm{L}2} &= m_{\mathrm{L}2}gr_{\mathrm{L}2,2} \\
                V_\mathrm{F} &= m_\mathrm{F}gr_{\mathrm{F},2} \\
                \begin{split}
                        V_\mathrm{S} &= \frac{1}{2} k_1 \big( u_{\alpha_1} + \alpha_{01} - \alpha_1 \big)^2 + \\&\frac{1}{2} k_2 \big( u_{\alpha_2} + \alpha_{02} -\alpha_1 - \alpha_2 \big)^2
                \end{split}
        \end{align}
\end{subequations}
Note that the segmented monopod model has series elastic actuation. Thus, $V_\mathrm{S}$ contains the motor positions $u_{\alpha_1}$ and~$u_{\alpha_2}$. The kinetic energy of each part is given by
\begin{subequations}
        \begin{align}
                T_\mathrm{B} &= \frac{1}{2} \big( m_\mathrm{B} \dot{x}^2 + m_\mathrm{B} \dot{z}^2 + j_\mathrm{B} \dot{\phi}^2 \big) \\
                T_{\mathrm{L}1} &= \frac{1}{2} \Bigg( \bigg(\frac{d\mathbf{r}_{\mathrm{L}1}}{dt}\bigg)\T m_{\mathrm{L}1} \frac{d\mathbf{r}_{\mathrm{L}1}}{dt} \Bigg) + \frac{1}{2} j_{\mathrm{L}1} (\dot{\phi}+\dot{\alpha_1})^2 \\
                T_{\mathrm{L}2} &= \frac{1}{2} \Bigg( \bigg(\frac{d\mathbf{r}_{\mathrm{L}2}}{dt}\bigg)\T m_{\mathrm{L}2} \frac{d\mathbf{r}_{\mathrm{L}2}}{dt} \Bigg) + \frac{1}{2} j_{\mathrm{L}2} (\dot{\phi}+\dot{\alpha_1}+\dot{\alpha_2})^2 \\
                T_{\mathrm{F}} &= \frac{1}{2} \Bigg( \bigg(\frac{d\mathbf{r}_{\mathrm{F}}}{dt}\bigg)\T m_{\mathrm{F}} \frac{d\mathbf{r}_{\mathrm{F}}}{dt} \Bigg)
        \end{align}
\end{subequations}
We repeat to compute \eqref{eq:Lagrangian}, \eqref{eq:LagrEoM}, and \eqref{eq:EoMwoCont} with the new values and receive the EoM for the segmented monopod. We define the constraints for each phase:
\begin{subequations}
        \begin{align}
                \mathbf{g}_1 &= 0 \\
                \mathbf{g}_2 &= \mathbf{r}_\mathrm{F}
        \end{align}
\end{subequations}
Table \ref{tab:phasePrism} contains the numbering of phases and the event functions, which are equal to the ones of the prismatic monopod model. We define the event functions
\begin{subequations}
        \begin{align}
                e_1^2(\x) &= - r_{\mathrm{F},2} \\
                e_2^1(\x) &= - \uplambda_{2,2}
        \end{align}
        \label{eq:eventSegm}
\end{subequations}
where $\boldsymbol{\uplambda}_2$ are the contact forces (Eq. \eqref{eq:contact}). If an event function crosses zero in positive direction, the event is triggered. The projection from phase $i$ to $j$ in Eq. \eqref{eq:discrete} holds true.

\begin{table}
\centering
    \begin{tabular}{l l c c}
        
        \thickhline
        Parameter &  & Value & Unit\\
        \hline
        total mass & $m_0$ & 1 & $\cdot$ \\
        zero length & $l_0$ & 0.24 & $\cdot$ \\
        gravitational constant & $g$ & 1 & $\cdot$ \\
        mass main body & $m_\mathrm{B}$ & 0.8399 & $m_0$ \\
        mass upper leg & $m_{\mathrm{L}1}$ & 0.0821 & $m_0$ \\
        mass lower leg & $m_{\mathrm{L}2}$ & 0.0780 & $m_0$ \\
        mass foot & $m_\mathrm{F}$ & $1\cdot10^{-10}$ & $m_0$ \\
        inertia main body & $j_\mathrm{B}$ & 0.0146 & $m_0 l_0^2$ \\
        inertia upper leg & $j_{\mathrm{L}1}$ & 0.0044 & $m_0 l_0^2$ \\
        inertia lower leg & $j_{\mathrm{L}2}$ & $1.759\cdot10^{-4}$ & $m_0 l_0^2$ \\
        stiffness spring joint 1 & $k_1$ & 1.2695 & $m_0 g l_0/\text{rad}$ \\
        damping spring joint 1 & $d_1$ & 0.1390 & $m_0 \sqrt{l_0^3 g^3}/\text{rad}$ \\
        angle uncompressed spring 1 & $\alpha_{01}$ & & rad \\
        stiffness spring joint 2 & $k_2$ & 1.3340 & $m_0 g l_0/\text{rad}$ \\
        damping spring joint 2 & $d_2$ & 0.0360 & $m_0 \sqrt{l_0^3 g^3}/\text{rad}$ \\
        angle uncompressed spring 2 & $\alpha_{02}$ & 0.35 & rad \\
        upper leg length & $l_1$ & 0.5 & $l_0$ \\
        lower leg length & $l_2$ & 0.5 & $l_0$ \\
        \makecell[l]{horizontal position joint 1 \\ w.r.t. CoM of main body} & $x_J$ & 0.0 & $l_0$ \\
        \makecell[l]{vertical position joint 1 \\ w.r.t. CoM of main body} & $z_J$ & 0.0 & $l_0$ \\
        \makecell[l]{horizontal position CoM of \\ upper leg w.r.t. joint 1} & $x_{L1}$ & 0.0 & $l_0$ \\
        \makecell[l]{vertical position CoM of \\ upper leg w.r.t. joint 1} & $z_{L1}$ & -0.1375 & $l_0$ \\
        \makecell[l]{horizontal position CoM of \\ lower leg w.r.t. joint 2} & $x_{L2}$ & -0.0833 & $l_0$ \\
        \makecell[l]{vertical position CoM of \\ upper leg w.r.t. joint 2} & $x_{L2}$ & -0.1667 & $l_0$ \\
        \thickhline
        
    \end{tabular}
    \caption{Parameters of the Segmented Monopod Model.}
    \label{tab:parSegm}
\end{table}

\subsection{Sagittal Quadruped}
To cover systems with increasing complexity we apply a sagittal quadruped model with $n_\mathrm{q} = 7$. The model can be seen as extension of the segmented monopod, since both hind and front legs share the same properties as the leg of the segmented monopod model. The general coordinates of the system are $\mathbf{q} = [x,z,\phi,\alpha_{\mathrm{hu}},\alpha_{\mathrm{hl}},\alpha_{\mathrm{fu}},\alpha_{\mathrm{fl}}]\T$. The motors are mounted in series with the springs with motor positions $\mathbf{u} = [u_\mathrm{hu},u_\mathrm{hl},u_\mathrm{fu},u_\mathrm{fl}]$. The model structure is depicted in Figure~\ref{fig:SagittalQuadruped}. Table~\ref{tab:parQuadru} contains the values of all parameters of the model.\par

\begin{figure}
    \label{fig:SagittalQuadruped}
    \centering
    \resizebox{0.7\linewidth}{!}{\input{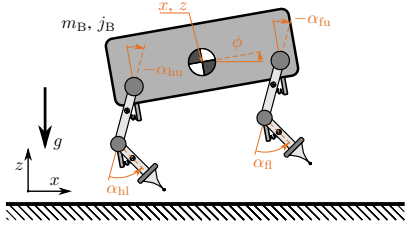}}
    \rev{\caption{The sagittal quadruped model contains a main body with mass $m_\mathrm{B}$ and inertia $j_\mathrm{B}$, and two legs (hind and front) consisting of an upper and lower leg, as well as a foot as point mass. In each leg there are two radial springs with stiffness and damping at each the hip and the knee joint. The configuration of the system is described by $\mathbf{q} = [x\,z\,\phi\,\alpha_{\mathrm{hu}}\,\alpha_{\mathrm{hl}}\,\alpha_{\mathrm{fu}}\,\alpha_{\mathrm{fl}}]\T$.}}
\end{figure}

The model consists of seven parts: base~$\mathrm{B}$, hind upper leg~$\mathrm{hu}$, hind lower leg~$\mathrm{hl}$, hind foot~$\mathrm{hf}$, front upper leg~$\mathrm{fu}$, front lower leg~$\mathrm{fl}$, and front foot~$\mathrm{ff}$. In the following, we define the position vectors of the system:
\begin{subequations}
        \begin{align}
                \mathbf{r}_\mathrm{B} &= [x,z]\T \\
                \mathbf{r}_\mathrm{hu} &= \begin{bmatrix}
                        x + x_\mathrm{Jh}c_\phi - z_\mathrm{J}s_\phi + x_\mathrm{u}c_{\phi\alpha_\mathrm{hu}} - z_\mathrm{u}s_{\phi\alpha_\mathrm{hu}} \\
                        z + x_\mathrm{Jh}s_\phi + z_\mathrm{J}c_\phi + x_\mathrm{u}s_{\phi\alpha_\mathrm{hu}} + z_\mathrm{u}c_{\phi\alpha_\mathrm{hu}}
                \end{bmatrix} \\
                \mathbf{r}_\mathrm{hl} &= \begin{bmatrix}
                        x + x_\mathrm{Jh}c_\phi - z_\mathrm{J}s_\phi + l_\mathrm{u}s_{\mathrm{\phi\alpha_\mathrm{hu}}}+ \\+ x_\mathrm{l}c_{\phi\alpha_\mathrm{hu}\alpha_\mathrm{hl}} - z_\mathrm{l}s_{\phi\alpha_\mathrm{hu}\alpha_\mathrm{hl}}; \\
                        z + x_\mathrm{Jh}s_\phi + z_\mathrm{J}c_\phi - l_\mathrm{u}c_{\phi\alpha_\mathrm{hu}} + \\+x_\mathrm{l}s_{\phi\alpha_\mathrm{hu}\alpha_\mathrm{hl}} + z_\mathrm{l}c_{\phi\alpha_\mathrm{hu}\alpha_\mathrm{hl}}
                \end{bmatrix} \\
                \mathbf{r}_\mathrm{hf} &= \begin{bmatrix}
                        x + x_\mathrm{Jh}c_\phi - z_\mathrm{J}s_\phi + l_\mathrm{u}s_{\phi\alpha_\mathrm{hu}} + l_\mathrm{l}s_{\phi\alpha_\mathrm{hu}\alpha_\mathrm{hl}} \\
                        z + x_\mathrm{Jh}s_\phi + z_\mathrm{J}c_\phi - l_\mathrm{u}c_{\phi\alpha_\mathrm{hu}} - l_\mathrm{l}c_{\phi\alpha_\mathrm{hu}\alpha_\mathrm{hl}}
                \end{bmatrix} \\
                \mathbf{r}_\mathrm{fu} &= \begin{bmatrix}
                        x + x_\mathrm{Jf}c_\phi - z_\mathrm{J}s_\phi + x_\mathrm{u}c_{\phi\alpha_\mathrm{fu}} - z_\mathrm{u}s_{\phi\alpha_\mathrm{fu}} \\
                        z + x_\mathrm{Jf}s_\phi + z_\mathrm{J}c_\phi + x_\mathrm{u}s_{\phi\alpha_\mathrm{fu}} + z_\mathrm{u}c_{\phi\alpha_\mathrm{fu}}
                \end{bmatrix} \\
                \mathbf{r}_\mathrm{fl} &= \begin{bmatrix}
                        x + x_\mathrm{Jf}c_\phi - z_\mathrm{J}s_\phi + l_\mathrm{u}s_{\mathrm{\phi\alpha_\mathrm{fu}}}+ \\+ x_\mathrm{l}c_{\phi\alpha_\mathrm{fu}\alpha_\mathrm{fl}} - z_\mathrm{l}s_{\phi\alpha_\mathrm{fu}\alpha_\mathrm{fl}}; \\
                        z + x_\mathrm{Jf}s_\phi + z_\mathrm{J}c_\phi - l_\mathrm{u}c_{\phi\alpha_\mathrm{fu}} + \\+x_\mathrm{l}s_{\phi\alpha_\mathrm{fu}\alpha_\mathrm{fl}} + z_\mathrm{l}c_{\phi\alpha_\mathrm{fu}\alpha_\mathrm{fl}}
                \end{bmatrix} \\
                \mathbf{r}_\mathrm{ff} &= \begin{bmatrix}
                        x + x_\mathrm{Jf}c_\phi - z_\mathrm{J}s_\phi + l_\mathrm{u}s_{\phi\alpha_\mathrm{fu}} + l_\mathrm{l}s_{\phi\alpha_\mathrm{fu}\alpha_\mathrm{fl}} \\
                        z + x_\mathrm{Jf}s_\phi + z_\mathrm{J}c_\phi - l_\mathrm{u}c_{\phi\alpha_\mathrm{fu}} - l_\mathrm{l}c_{\phi\alpha_\mathrm{fu}\alpha_\mathrm{fl}}
                \end{bmatrix}
        \end{align}
\end{subequations}
where $c_{ij}$ and $s_{ij}$ denote $\cos(i+j)$ and $\sin(i+j)$, respectively. We utilize the position vectors to define the potential energy of each part $k$:
\begin{subequations}
        \begin{align}
                V_k &= m_k g r_{k,2} \\
                \begin{split}
                        V_\mathrm{S} &= \frac{1}{2}k\big( (u_\mathrm{hu} + \alpha_{0\mathrm{hu}} - \alpha_\mathrm{hu})^2 \\
                        &+(u_\mathrm{hl} + \alpha_{0\mathrm{hl}} - \alpha_\mathrm{hu}-\alpha_\mathrm{hl})^2\\
                        &+(u_\mathrm{fu} + \alpha_{0\mathrm{fu}} - \alpha_\mathrm{fu})^2 \\
                        &+(u_\mathrm{fl} + \alpha_{0\mathrm{fl}} - \alpha_\mathrm{fu}-\alpha_\mathrm{fl})^2\big)
                \end{split}
        \end{align}
\end{subequations}
The sagittal quadruped model is equipped with series elastic actuators. Therefore, $\bu$ is part of the potential energy of the springs. We define the kinetic energy of each part $k$
\begin{equation}
        T_k = \frac{1}{2}\Bigg( \bigg(\frac{d\mathbf{r}_k}{dt}\bigg)\T m_k \frac{d\mathbf{r}_k}{dt} \Bigg) + \frac{1}{2} j_k \omega_k^2
\end{equation}
where $\omega_k$ is the rotational velocity of the part $k$. Note that the feet are point masses and do not feature rotational velocity. With Equations \eqref{eq:Lagrangian}, \eqref{eq:LagrEoM}, and \eqref{eq:EoMwoCont} we receive the EoM of the sagittal quadruped. Table \ref{tab:phaseQuadru} shows the numbering of the phases. The constraints of each phase are defined in the following
\begin{subequations}
        \begin{align}
                \mathbf{g}_1 &= 0\\
                \mathbf{g}_2 &= \mathbf{r}_\mathrm{hf}\\
                \mathbf{g}_3 &= \mathbf{r}_\mathrm{ff}\\
                \mathbf{g}_4 &= \begin{bmatrix}
                        \mathbf{r}_\mathrm{hf}\\
                        \mathbf{r}_\mathrm{ff}
                \end{bmatrix}
        \end{align}
\end{subequations}
The event functions for the sagittal quadruped model are similar to Eq. \eqref{eq:eventSegm}. For lift-offs we observe the respective $z$-direction of the contact forces. For touch-downs we observe the height of the feet over the ground.

\begin{table}
        \centering
        \begin{tabular}{c l}
                \thickhline
                \textbf{no.} & \textbf{phase} \\
                \hline
                1 & flight \\
                2 & single-support hind leg \\
                3 & single-support front leg \\
                4 & double-support \\
                \thickhline
        \end{tabular}
        \caption{Phases of the sagittal quadruped model.}
        \label{tab:phaseQuadru}
\end{table}

\begin{table}
\centering
    \begin{tabular}{l l c c}
        
        \thickhline
        Parameter &  & Value & Unit\\
        \hline
        total mass & $m_0$ & 1 & $\cdot$ \\
        zero length & $l_0$ & 0.16 & $\cdot$ \\
        gravitational constant & $g$ & 1 & $\cdot$ \\
        mass main body & $m_\mathrm{B}$ & 0.8797 & $m_0$ \\
        mass upper leg & $m_{\mathrm{u}}$ & 0.0430 & $m_0$ \\
        mass lower leg & $m_{\mathrm{l}}$ & 0.0172 & $m_0$ \\
        mass foot & $m_\mathrm{F}$ & 0 & $m_0$ \\
        inertia main body & $j_\mathrm{B}$ & 0.7744 & $m_0 l_0^2$ \\
        inertia upper leg & $j_{\mathrm{r}}$ & $8.9826\cdot10^{-4}$ & $m_0 l_0^2$ \\
        inertia lower leg & $j_{\mathrm{l}}$ & $3.6201\cdot10^{-4}$ & $m_0 l_0^2$ \\
        stiffness rotational joints & $k$ & 0.9472 & $m_0 g l_0/\text{rad}$ \\
        damping rotational joints & $d$ & 0.038 & $m_0 \sqrt{l_0^3 g^3}/\text{rad}$ \\
        angle uncompr. spring (up) & $\alpha_{0\mathrm{u}}$ & -0.4 & rad \\
        angle uncompr. spring (low) & $\alpha_{0\mathrm{l}}$ & 0.4 & rad \\
        upper leg length & $l_\mathrm{u}$ & 0.5 & $l_0$ \\
        lower leg length & $l_\mathrm{l}$ & 0.5 & $l_0$ \\
        \makecell[l]{horizontal pos. joint hu \\ w.r.t. CoM of main body} & $x_\mathrm{Jh}$ & -0.9375 & $l_0$ \\
        \makecell[l]{horizontal pos. joint fu \\ w.r.t. CoM of main body} & $x_\mathrm{Jf}$ & 0.9375 & $l_0$ \\
        \makecell[l]{vertical pos. joints (up) \\ w.r.t. CoM of main body} & $z_\mathrm{J}$ & -0.2012 & $l_0$ \\
        \makecell[l]{horizontal position CoM of \\ upper leg w.r.t. upper joint} & $x_{\mathrm{u}}$ & 0.0010 & $l_0$ \\
        \makecell[l]{vertical position CoM of \\ upper leg w.r.t. upper joint} & $z_{\mathrm{u}}$ & -0.2110 & $l_0$ \\
        \makecell[l]{horizontal position CoM of \\ lower leg w.r.t. lower joint} & $x_{\mathrm{l}}$ & 0 & $l_0$ \\
        \makecell[l]{vertical position CoM of \\ lower leg w.r.t. lower joint} & $z_{\mathrm{l}}$ & -0.2398 & $l_0$ \\
        \thickhline
        
    \end{tabular}
    \caption{Parameters of the Sagittal Quadruped Model.}
    \label{tab:parQuadru}
\end{table}

\section{Root-Search with Direct Collocation}
\label{sec:root}
We apply direct collocation to (main document, Eq. (4)). For this, we use CasADi \cite{Andersson2019}, a tool for nonlinear optimization and algorithmic differentiation. The function \verb|nlpsol| allows to set the solver \verb|IPOPT|, as well as the struct \verb|nlp|, that we want to solve. This struct contains a decision variable $\mathbf{a}$, a vector containing the residual parameters, a cost function, which here is set to zero, and the constraint vector. In the following, we describe the content of the constraint vector.\par
For each model we set up a \verb|FlowMap| (containing the continuous flow), a \verb|JumpSet| (containing the event functions), and a \verb|JumpMap| (containing the discrete mapping between phases). This is done with the \verb|SX| symbolics of CasADi. The first constraint we set is the \textit{derivation violation} of the Hermite-Simpson approximation for the direct collocation framework. We apply the compressed form of the collocation constraints and compare the approximated dynamics at the center of a grid interval to the actual system dynamics. This is repeated for all intervals. Note that discrete jumps between phases have to be considered using the \verb|JumpMap|.\par
Next, we consider the \textit{periodicity constraints}, which we set for each desired periodic state. The \textit{event search constraints} are used to guarantee that at each end of a phase there is the correct event function fulfilled. The last constraint is the \textit{operating point constraint}, which sets either an average speed or a certain energy level as constraint for a gait.\par
All constraints are gathered in \verb|nlpsol| and the lower and upper bound for the constraints is set to zero. The decision variable $\mathbf{a}$ contains the initial states of the gait $\mathbf{x}_0$, the energy injection parameter $\gamma$, the time points of the events $\mathbf{t}_\mathrm{event}$, and the values of the states at each grid point $\mathbf{x}_\mathrm{grid}$. To receive an initial guess for these values, we try different initial states of a gait until the correct sequence of phases is reached in simulation. The NLP solver gives back an approximated quasi-passive, periodic gait with adapted energy injection parameter~$\gamma$.\par

Tables \ref{tab:PrismForward}-\ref{tab:QuadruForward} contain the initial guesses and results after root-search for the initial states of the quasi-passive gait for each of the three exemplary models.

\begin{table}
\centering
    \begin{tabular}{ 
         l l c  c  c  }
        
        \thickhline
        States &  & \makecell{\, Initial \,\\ Guess} & Result & Unit\\
        \hline
        horizontal position & $x$ & 0  & 0 & $l_0$ \\
        vertical position & $z$ & 1.0000  & 0.9713 & $l_0$ \\
        absolute rotation & $\phi$ & 0  & 0.0676 & rad \\
        angle rotational joint & $\alpha$ & 0  & 0.1723 & rad \\
        length of leg & $l$ & 1.0000  & 0.9999 & $l_0$ \\
        horizontal velocity & $\Dot{x}$ & 0 & 0.3835 & $\sqrt{l_0 g}$ \\
        vertical velocity & $\Dot{z}$ & -0.5500  & -0.5740 & $\sqrt{l_0 g}$ \\
        abs. rot. velocity & $\Dot{\phi}$ & 0  & 0.0082 & rad$/ \sqrt{l_0 g}$ \\
        rot. velocity joint & $\Dot{\alpha}$ & 0  & -0.2443 & rad$/ \sqrt{l_0 g}$ \\
        velocity leg length & $\Dot{l}$ & -0.5500 & -0.6487 & $\sqrt{l_0 g}$ \\
        \hline
        virtual energy injection & $\gamma$ & 0.0100 & 0.4466 & $\cdot$ \\
        \thickhline
        
    \end{tabular}
    \caption{Hopping forward gait of the prismatic monopod model with virtual energy injection.}
    \label{tab:PrismForward}
\end{table}

\begin{table}
\centering
    \begin{tabular}{ 
        l l c  c  c }
        
        \thickhline
        States &  & \makecell{\, Initial \,\\ Guess}  & Result & Unit\\
        \hline
        horizontal position & $x$ & 0  & 0 & $l_0$ \\
        vertical position & $z$ & 0.9613  & 0.8995 & $l_0$ \\
        absolute rotation & $\phi$ & -0.1234  & 0.2877 & rad \\
        angle joint 1 & $\alpha_1$ & -0.2000  & -0.1806 & rad \\
        angle joint 2 & $\alpha_2$ & 0.5500  & 0.5283 & rad \\
        horizontal velocity & $\Dot{x}$ & 0.0364 & 0.4652 & $\sqrt{l_0 g}$ \\
        vertical velocity & $\Dot{z}$ & -0.7500  & -0.4227 & $\sqrt{l_0 g}$ \\
        abs. rot. velocity & $\Dot{\phi}$ & 0  & 0.4881 & rad$/ \sqrt{l_0 g}$ \\
        rot. vel. joint 1 & $\Dot{\alpha}_1$ & -2.8275  & -2.9334 & rad$/ \sqrt{l_0 g}$ \\
        rot. vel. joint 2 & $\Dot{\alpha}_2$ & 5.5042 & 4.3100 & rad$/ \sqrt{l_0 g}$ \\
        \hline
        virtual energy injection & $\gamma$ & 0.0100  & 0.8356 & $\cdot$ \\
        angle uncompr. spring 1 & $\alpha_{01}$ & -0.2000  & -0.1761 & rad \\
        \thickhline
        
    \end{tabular}
    \caption{Hopping forward gait of the segmented monopod model with virtual energy injection.}
    \label{tab:SegmForward}
\end{table}

\begin{table}
\centering
    \begin{tabular}{ 
        l l c  c  c }
        
        \thickhline
        States &  & \makecell{\, Initial \,\\ Guess}  & Result & Unit\\
        \hline
        horizontal position & $x$ & 0  & 0 & $l_0$ \\
        vertical position & $z$ & 1.2407  & 1.1248 & $l_0$ \\
        absolute rotation & $\phi$ & 0.1237  & 0.0015 & rad \\
        hind upper joint & $\alpha_\mathrm{hu}$ & -0.3656  & -0.3989 & rad \\
        hind lower joint & $\alpha_\mathrm{hl}$ & 0.7371  & 0.7943 & rad \\
        front upper joint & $\alpha_\mathrm{fu}$ & -0.3521  & -0.3657 & rad \\
        front lower joint & $\alpha_\mathrm{fl}$ & 0.7398  & 0.7997 & rad \\
        horizontal velocity & $\Dot{x}$ & 0.3173 & 0.4112 & $\sqrt{l_0 g}$ \\
        vertical velocity & $\Dot{z}$ & -1.2092  & -0.4870 & $\sqrt{l_0 g}$ \\
        abs. rot. velocity & $\Dot{\phi}$ & -0.4806  & 0.2990 & rad$/ \sqrt{l_0 g}$ \\
        rot. vel. hind upper & $\Dot{\alpha}_\mathrm{hu}$ & -1.7800  & -2.7939 & rad$/ \sqrt{l_0 g}$ \\
        rot. vel. hind lower & $\Dot{\alpha}_\mathrm{hl}$ & 4.3786 & 3.9662 & rad$/ \sqrt{l_0 g}$ \\
        rot. vel. front upper & $\Dot{\alpha}_\mathrm{fu}$ & -0.4504  & 0.2260 & rad$/ \sqrt{l_0 g}$ \\
        rot. vel. front lower & $\Dot{\alpha}_\mathrm{fl}$ & 0.1995 & -0.3944 & rad$/ \sqrt{l_0 g}$ \\
        \hline
        virtual energy injection & $\gamma$ & 0.0100  & 0.6910 & $\cdot$ \\
        \thickhline
        
    \end{tabular}
    \caption{Bounding gait of the sagittal quadruped model with virtual energy injection.}
    \label{tab:QuadruForward}
\end{table}

\section{Homotopic Continuation}
\label{sec:contin}
Algorithm \ref{alg:HomotopicContinuation} shows the implementation of homotopic continuation on the first-order optimality. We start with modifying the decision variable $\mathbf{a}$ given from direct collocation. Since we do not want to change the virtual energy injection parameter~$\gamma$, but the homotopy parameter~$\varepsilon$, we remove~$\gamma$ from~$\mathbf{a}$. Thus,~$\gamma$ stays constant over the continuation. To receive a non-zero actuation profile, we add the input parameters~$\bxi$ to the decision variable (line 1).\par
We set up the cost function~$c$ for the respective system and define CasADi functions for the homotopy map
\begin{equation}
\label{eq:HomotopyMap}
    \br(\bzeta, \varepsilon) := \frac{\partial L_{\bar{O}}}{\partial \bzeta}(\bzeta, \varepsilon) = 
    \begin{bmatrix}
        (\frac{\partial c}{\partial \ba})\T + (\frac{\partial \mathbf{h}}{\partial \ba})\T \blambda \\
        \mathbf{h}(\cdot)
    \end{bmatrix}.
\end{equation}
and its partial derivative
\begin{equation}
    \label{eq:R}
    \mathbf{R}(\boldsymbol{\zeta}, \varepsilon) = \frac{\partial \mathbf{r} \, (\boldsymbol{\zeta}, \varepsilon)}{\partial (\boldsymbol{\zeta}, \varepsilon)}.
\end{equation}
which is done by using the \verb|jacobian| function of CasADi. The initial value for the Lagrange multiplier $\blambda$ is computed as follows
\begin{equation}
        \label{eq:blambda}
        \blambda^* = - \Big(\frac{\partial \mathbf{h}}{\partial \ba}\Big)^{-\top} \Big|_{\ba = \ba^*} \Big(\frac{\partial c}{\partial \ba}\Big)\T \Big|_{\ba = \ba^*}
\end{equation}
This allows the computation of the initial tangent vector $\mathbf{p}^*$:
\begin{subequations}
    \label{eq:TangentVector}
    \begin{align}
        \mathbf{R}(\boldsymbol{\zeta}^*, \varepsilon^*) \cdot \mathbf{p} = \mathbf{0}, \\
        \Vert \mathbf{p} \Vert_2 = 1, \\
        \det \Biggl( 
        \begin{bmatrix}
            \mathbf{R}(\boldsymbol{\zeta}^*, \varepsilon^*) \\
            \mathbf{p}\T
        \end{bmatrix}    
        \Biggr) > 0.
    \end{align}
\end{subequations}
With the definition of the orientation~$d_\mathrm{hom}$ in direction of increasing~$\varepsilon$ (line 5) we can start the continuation. First, a predictor step in the direction of the tangent vector is done (line 8). This allows the computation of a new tangent vector~$\mathbf{p}_\mathrm{pred}$. Note that a step size adaption can be an useful implementation here \cite{allgower2003introduction}. A Newton corrector step (line 11) is applied as long as~$\br$ is larger as a given tolerance ($10^{-6}$). We repeat this procedure until $\varepsilon = 1$. To ensure that the result is a local minimum we check the second-order optimality condition as sufficient condition
\begin{equation}
    \label{eq:SecondOrdOpt}
    \mathbf{y}\T \, \Biggl( \frac{\partial^2 c}{\partial \ba^2} (\ba,\varepsilon) + \blambda\T \frac{\partial^2 \mathbf{h}}{\partial \ba^2} (\ba, \bar{O},\varepsilon)\Biggr) \, \mathbf{y} > 0,
\end{equation}

\begin{algorithm}
    \caption{Continuation on First-Order Optimality}
    \label{alg:HomotopicContinuation}
    \DontPrintSemicolon
    \KwIn{Quasi-passive gait $\x_0^*, t_{\text{event}}^*, \x_{\text{grid}}^*, \bar{O}, \epsilon^* = 0$, step size $\delta$, cost $c$}
    \KwOut{Actuated gait $\{\x_0, t_{\text{event}}, \x_{\text{grid}}, \boldsymbol{\xi}\}_{\text{solution}}, \varepsilon = 1$}
    remove $\gamma$ from $\mathbf{a}$, add $\boldsymbol{\xi}$ to $\mathbf{a}$\;
    $\boldsymbol{\psi}^* = (\boldsymbol{\zeta}^*, \varepsilon^*)$ with $\boldsymbol{\zeta}^* = (\mathbf{a}^*, \boldsymbol{\lambda}^*)$\;
    uniquely solve for $\boldsymbol{\lambda}^*$ \tcp*{(\ref{eq:blambda})} 
    compute initial tangent vector $\mathbf{p}^*$ \tcp*{(\ref{eq:TangentVector})}
    $d_{\text{hom}} \leftarrow$ sign$(\mathbf{p}^*(\text{end}))$\;
    $\boldsymbol{\psi} \leftarrow \boldsymbol{\psi}^*, \mathbf{p} \leftarrow \mathbf{p}^*, \varepsilon \leftarrow \varepsilon^*$\;
    \While{$\varepsilon < 1$}{
        $\boldsymbol{\psi}_{\text{pred}} \leftarrow \boldsymbol{\psi} + \delta d_{\text{hom}} \mathbf{p}$ \tcp*{predictor step}
        compute $\mathbf{p}_{\text{pred}}$ at $\boldsymbol{\psi}_{\text{pred}}$ \tcp*{(\ref{eq:TangentVector})}
        \While{max$(\mathbf{r}) >$ tolerance}{
            $\boldsymbol{\psi} \leftarrow \boldsymbol{\psi} - \begin{bmatrix}
                \mathbf{R}(\boldsymbol{\psi}) \\
                \mathbf{p}^T
            \end{bmatrix}^{-1}
            \cdot \begin{bmatrix}
                \mathbf{r}(\boldsymbol{\psi}) \\
                0
            \end{bmatrix}$ \tcp*{corrector step}
            \Return{$\boldsymbol{\psi}_{\text{corr}}, \mathbf{p}_{\text{corr}}$}
        }
        $\boldsymbol{\psi} \leftarrow \boldsymbol{\psi}_{\text{corr}}, \mathbf{p} \leftarrow \mathbf{p}_{\text{corr}}, \varepsilon \leftarrow \varepsilon_{\text{corr}}$\;
    }
    check second-order optimality condition of $\boldsymbol{\psi} = (\boldsymbol{\zeta}, 1)$ \tcp*{(\ref{eq:SecondOrdOpt})}
\end{algorithm}

After continuation, the decision variable contains the initial states $\x_0$ of the optimally actuated gait, the time points of events $\mathbf{t}_\mathrm{event}$, the values of the states at each grid point of the actuated gait $\x_\mathrm{grid}$, and the actuation parameters $\bxi$, which define the actuation profile of the gait.\par
In the following, the results for an optimally actuated gait of the segmented monopod are shown. This complements the two examples shown in the main document.

\subsection{Example: Actuated Gait of Segmented Monopod}
We start with the direct collocation approach to find quasi-passive gaits. We set the number of grid points to $N=18$ per phase. For the operating point constraint we set a hopping forward gait with average speed $\bar{\dot{x}} = 0.3 \sqrt{l_0g}$.
Figure~\ref{fig:SegmForward} depicts the resulting approximated trajectory of the states.\par
We apply the concept of homotopic continuation and derive an actuation profile for actuated gaits on this model. We choose the cost function $c = \boldsymbol{\xi}\T \boldsymbol{\xi}$.
The segmented monopod model uses series elastic actuation; thus the control inputs are the motor positions $u_{\alpha_1}$ and $u_{\alpha_2}$ at joint 1 and 2.\par
\begin{figure}
\begin{minipage}{0.05\linewidth}
    \rotatebox{90}{\footnotesize Input Motor Pos. \hspace{1.5cm} Continuous States \hspace{0.7cm}}
\end{minipage}
\begin{minipage}{0.95\linewidth}
    \pgfplotsset{footnotesize}
    \begin{tikzpicture}
        \begin{axis}[
            name=main plot,
            title={Segmented Monopod: Actuated Gait},
            xtick={0,.5,1,1.5,2},
            ytick={-.4,0,.4,.8,1.2},
            xmin=0,
            xmax=2,
            ymin=-0.6,
            ymax=1.3,
            legend pos=outer north east,
            grid = major,
            legend entries={$x$,$z$,$\phi$,$\alpha_1$,$\alpha_2$},
            width=7cm,
            height=3.5cm
        ]

            \addplot[tumBlue, line width=1.2pt] table [
                col sep=comma,
                mark=none
            ]{Data/Figure06_1/Figure06_1_x.csv};

            \addplot[tumOrange, line width=1.2pt] table [
                col sep=comma,
                mark=none
            ]{Data/Figure06_1/Figure06_1_z.csv};

            \addplot[tumYellow, line width=1.2pt] table [
                col sep=comma,
                mark=none
            ]{Data/Figure06_1/Figure06_1_phi.csv};

            \addplot[tumPink, line width=1.2pt] table [
                col sep=comma,
                mark=none
            ]{Data/Figure06_1/Figure06_1_gamma1.csv};

            \addplot[tumGreen, line width=1.2pt] table [
                col sep=comma,
                mark=none
            ]{Data/Figure06_1/Figure06_1_gamma2.csv};

            \path[name path=axis] (0,-10) -- (0,10);
            \path[name path=f] (1.43,-10) -- (1.43,10);
            \addplot[
                color=gray,
                fill opacity=0.2
            ] fill between[
                of=axis and f
            ];
            
        \end{axis}

        \begin{axis}[
            name=second plot,
            at={(main plot.below south west)},
            yshift=-0.1cm,
            anchor=north west,
            xtick={0,.5,1,1.5,2},
            ytick={-3,-2,-1,0,1,2,3,4},
            xmin=0,
            xmax=2,
            ymin=-3,
            ymax=4,
            legend pos=outer north east,
            grid = major,
            legend entries={$\Dot{x}$,$\Dot{z}$,$\Dot{\phi}$,$\Dot{\alpha}_1$,$\Dot{\alpha}_2$},
            width=7cm,
            height=3.5cm
        ]

            \addplot[tumBlue, line width=1.2pt] table [
                col sep=comma,
                mark=none
            ]{Data/Figure06_2/Figure06_2_dx.csv};

            \addplot[tumOrange, line width=1.2pt] table [
                col sep=comma,
                mark=none
            ]{Data/Figure06_2/Figure06_2_dz.csv};

            \addplot[tumYellow, line width=1.2pt] table [
                col sep=comma,
                mark=none
            ]{Data/Figure06_2/Figure06_2_dphi.csv};

            \addplot[tumPink, line width=1.2pt] table [
                col sep=comma,
                mark=none
            ]{Data/Figure06_2/Figure06_2_dgamma1.csv};

            \addplot[tumGreen, line width=1.2pt] table [
                col sep=comma,
                mark=none
            ]{Data/Figure06_2/Figure06_2_dgamma2.csv};

            \path[name path=axis] (0,-10) -- (0,10);
            \path[name path=f] (1.43,-10) -- (1.43,10);
            \addplot[
                color=gray,
                fill opacity=0.2
            ] fill between[
                of=axis and f
            ];
            
        \end{axis}

        \begin{axis}[
            at={(second plot.below south west)},
            yshift=-0.1cm,
            anchor=north west,
            xlabel={Time $[\sqrt{l_0 / g}]$},
            xtick={0,.5,1,1.5,2},
            ytick={-.3,-.2,-.1,0,.1,.2},
            xmin=0,
            xmax=2,
            ymin=-.3,
            ymax=.2,
            legend pos=outer north east,
            grid = major,
            legend entries={$u_{\alpha_1}$,$u_{\alpha_2}$},
            width=7cm,
            height=3cm
        ]

            \addplot+[jump mark left, tumPink, line width=1.2pt, mark size=1pt] table [
                col sep=comma
            ]{Data/Figure06_3/Figure06_3_ugamma1.csv};

            \addplot+[jump mark left, tumGreen, line width=1.2pt, mark size=1pt, mark=*] table [
                col sep=comma
            ]{Data/Figure06_3/Figure06_3_ugamma2.csv};

            \path[name path=axis] (0,-10) -- (0,10);
            \path[name path=f] (1.43,-10) -- (1.43,10);
            \addplot[
                color=gray,
                fill opacity=0.2
            ] fill between[
                of=axis and f
            ];
            
        \end{axis}
    \end{tikzpicture}
    \end{minipage}
    \caption{Actuated hopping forward gait of the segmented monopod as result of homotopic continuation. The cost function is $c = \boldsymbol{\xi}\T \boldsymbol{\xi}$. The gray area marks the stance phase. On the bottom, the motor trajectory is depicted. Note that the segmented monopod employs series elastic actuation.}
    \label{fig:ActSegm}
\end{figure}
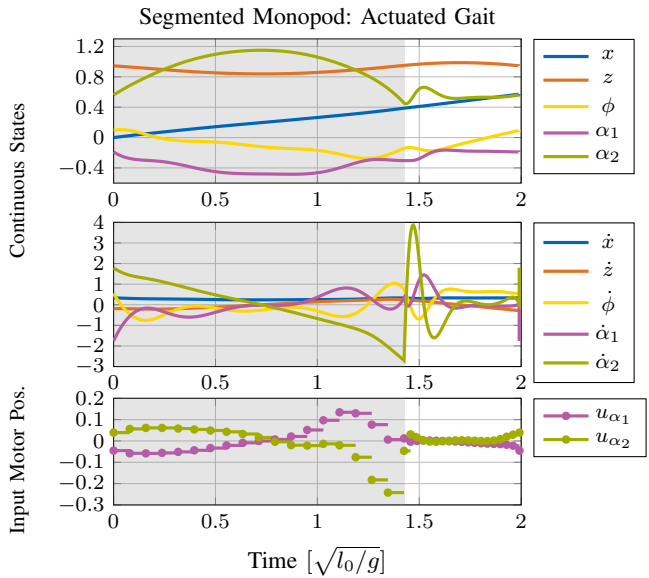

We apply homotopic continuation on the quasi-passive gait and find an actuated gait without virtual energy injection. The obtained motor trajectory is shown in Figure~\ref{fig:ActSegm}. It is observable that during the stance phase, the motor supports the bending and pushing-off motion of the leg. Overall, the applied methods lead to an optimally actuated gait, i.e., the computed input is locally minimal with respect to the defined cost function, which is verified with \eqref{eq:SecondOrdOpt}. This allows the system to move forward efficiently.

\begin{figure}
\begin{minipage}{0.05\linewidth}
    \centering
    \rotatebox{90}{\footnotesize Continuous States}
    \vspace{0.75cm}
\end{minipage}
\begin{minipage}{0.95\linewidth}
    \pgfplotsset{footnotesize}
    \begin{tikzpicture}
        \begin{axis}[
            name=main plot,
            title={Segmented Monopod: Quasi-Passive Gait},
            xtick={0,.4,.8,1.2,1.6,2,2.4},
            ytick={-.8,-.4,0,.4,.8,1.2,1.6},
            xmin=0,
            xmax=2.4,
            ymin=-0.8,
            ymax=1.6,
            legend pos=outer north east,
            grid = major,
            legend entries={$x$,$z$,$\phi$,$\alpha_1$,$\alpha_2$},
            width=7cm,
            height=3.5cm
        ]

            \addplot[tumBlue, line width=1.2pt] table [
                col sep=comma,
                mark=none
            ]{Data/Figure04_1/Figure04_1_x.csv};

            \addplot[tumOrange, line width=1.2pt] table [
                col sep=comma,
                mark=none
            ]{Data/Figure04_1/Figure04_1_z.csv};

            \addplot[tumYellow, line width=1.2pt] table [
                col sep=comma,
                mark=none
            ]{Data/Figure04_1/Figure04_1_phi.csv};

            \addplot[tumPink, line width=1.2pt] table [
                col sep=comma,
                mark=none
            ]{Data/Figure04_1/Figure04_1_gamma1.csv};

            \addplot[tumGreen, line width=1.2pt] table [
                col sep=comma,
                mark=none
            ]{Data/Figure04_1/Figure04_1_gamma2.csv};

            \path[name path=axis] (0,-10) -- (0,10);
            \path[name path=f] (1.31,-10) -- (1.31,10);
            \addplot[
                color=gray,
                fill opacity=0.2
            ] fill between[
                of=axis and f
            ];
            
        \end{axis}

        \begin{axis}[
            at={(main plot.below south west)},
            yshift=-0.1cm,
            anchor=north west,
            xlabel={Time $[\sqrt{l_0 / g}]$},
            xtick={0,.4,.8,1.2,1.6,2,2.4},
            ytick={-3,-2,-1,0,1,2,3,4,5},
            xmin=0,
            xmax=2.4,
            ymin=-3,
            ymax=5,
            legend pos=outer north east,
            grid = major,
            legend entries={$\Dot{x}$,$\Dot{z}$,$\Dot{\phi}$,$\Dot{\alpha}_1$,$\Dot{\alpha}_2$},
            width=7cm,
            height=3.5cm
        ]

            \addplot[tumBlue, line width=1.2pt] table [
                col sep=comma,
                mark=none
            ]{Data/Figure04_2/Figure04_2_dx.csv};

            \addplot[tumOrange, line width=1.2pt] table [
                col sep=comma,
                mark=none
            ]{Data/Figure04_2/Figure04_2_dz.csv};

            \addplot[tumYellow, line width=1.2pt] table [
                col sep=comma,
                mark=none
            ]{Data/Figure04_2/Figure04_2_dphi.csv};

            \addplot[tumPink, line width=1.2pt] table [
                col sep=comma,
                mark=none
            ]{Data/Figure04_2/Figure04_2_dgamma1.csv};

            \addplot[tumGreen, line width=1.2pt] table [
                col sep=comma,
                mark=none
            ]{Data/Figure04_2/Figure04_2_dgamma2.csv};

            \path[name path=axis] (0,-10) -- (0,10);
            \path[name path=f] (1.31,-10) -- (1.31,10);
            \addplot[
                color=gray,
                fill opacity=0.2
            ] fill between[
                of=axis and f
            ];
            
        \end{axis}
    \end{tikzpicture}
    \end{minipage}
    \caption{The continuous states of the segmented monopod model for a hopping forward gait with virtual energy injection. The operating point constraint is the average speed $\Bar{\Dot{x}} = 0.3\sqrt{l_0g}$. The gray area marks the stance phase. The springs are compressed over the stance phase and swing back to their uncompressed position in the flight phase. All states return to their initial values with the impact at the end of the stride.}
    \label{fig:SegmForward}
\end{figure}
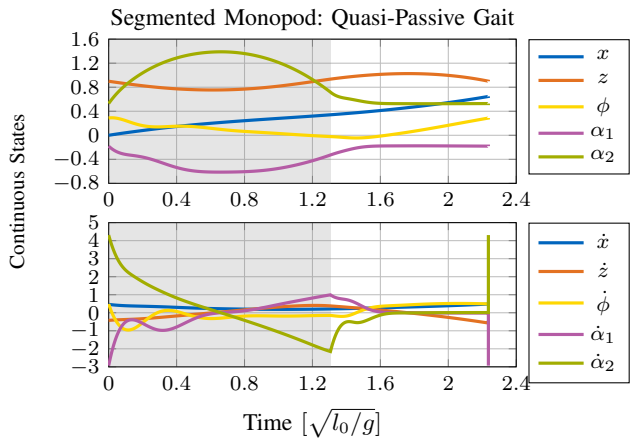

\end{document}